\documentclass[sigconf,nonacm]{acmart}

\usepackage{popets}
\usepackage{paralist}
\usepackage{amsmath}
\usepackage{booktabs}
\usepackage{pifont}
\usepackage{array}
\usepackage{graphicx}
\usepackage{stfloats}
\usepackage{placeins}
\usepackage{multirow}
\usepackage{tabularx,enumitem}
\usepackage{balance}
\usepackage{makecell}

\usepackage{multirow}
\usepackage{tabularx,enumitem}
\usepackage{balance}
\usepackage{makecell}

\setcopyright{popets}
\copyrightyear{YYYY}

\acmYear{YYYY}
\acmVolume{YYYY}
\acmNumber{X}
\acmDOI{XXXXXXX.XXXXXXX}
\acmISBN{}
\acmConference{Proceedings on Privacy Enhancing Technologies}
\newlength{\saveparindent}
\newlength{\saveparskip}
\newcounter{ctr}

\begin{document}

\newcommand{\cmark}{\ding{51}} 
\newcommand{\xmark}{\ding{55}} %

\title{Benchmarking LLM Privacy Recognition for Social Robot Decision Making}


\author{Dakota Sullivan*}
\orcid{0000-0002-4901-6375}
\affiliation{
  \institution{University of Wisconsin - Madison}
  \city{Madison}
  \state{Wisconsin}
  \country{USA}}
\email{dsullivan8@wisc.edu}

\author{Shirley Zhang*}
\orcid{0009-0000-5810-1347}
\affiliation{
  \institution{University of Wisconsin - Madison}
  \city{Madison}
  \state{Wisconsin}
  \country{USA}}
\email{hzhang664@wisc.edu}

\author{Jennica Li}
\orcid{0000-0002-4697-0948}
\affiliation{
  \institution{University of Wisconsin - Madison}
  \city{Madison}
  \state{Wisconsin}
  \country{USA}}
\email{jennica.li@wisc.edu}

\author{Heather Kirkorian}
\orcid{0000-0002-7990-7777}
\affiliation{
  \institution{University of Wisconsin - Madison}
  \city{Madison}
  \state{Wisconsin}
  \country{USA}}
\email{kirkorian@wisc.edu}

\author{Bilge Mutlu}
\orcid{0000-0002-9456-1495}
\affiliation{
  \institution{University of Wisconsin - Madison}
  \city{Madison}
  \state{Wisconsin}
  \country{USA}}
\email{bilge@wisc.edu}

\author{Kassem Fawaz}
\orcid{0000-0002-4609-7691}
\affiliation{
  \institution{University of Wisconsin - Madison}
  \city{Madison}
  \state{Wisconsin}
  \country{USA}}
\email{kfawaz@wisc.edu}


\renewcommand{\shortauthors}{Sullivan et al.}

\begin{abstract}

While robots have previously utilized rule-based systems or probabilistic models for user interaction, the rapid evolution of large language models (LLMs) presents new opportunities to develop LLM-powered robots for enhanced human-robot interaction (HRI). To fully realize these capabilities, however, robots need to collect data such as audio, fine-grained images, video, and locations. As a result, LLMs often process sensitive personal information, particularly within private environments, such as homes. Given the tension between utility and privacy risks, evaluating how current LLMs manage sensitive data is critical. Specifically, we aim to explore the extent to which out-of-the-box LLMs are privacy-aware in the context of household robots. In this work, we present a set of privacy-relevant scenarios developed using the Contextual Integrity (CI) framework. We first surveyed users’ privacy preferences regarding in-home robot behaviors and then examined how their privacy orientations affected their choices of these behaviors ($N = 450$). We then provided the same set of scenarios and questions to state-of-the-art LLMs ($N = 10$) and found that the agreement between humans and LLMs was generally low. To further investigate the capabilities of LLMs as potential privacy controllers, we implemented four additional prompting strategies and compared their results. We discuss the performance of the evaluated models as well as the implications and potential of AI privacy awareness in human-robot interaction.




\end{abstract}

\keywords{privacy, large language models, human-robot interaction}

\maketitle
\def\thefootnote{*}\footnotetext{Both authors contributed equally to this research.}\def\thefootnote{\arabic{footnote}}

\section{Introduction}
As robots become more competent and provide greater utility, we increasingly rely on them to perform tasks in private locations such as homes~\cite{nysofa_elliq_initiative_2025} and care settings ~\cite{Jack2024_PARIS, TheSun2025_NuraBot}. As early as 2014, more than 3,000 Paro therapeutic robots had been sold across 30 countries~\cite{AIST_2013_Paro}, and by 2024, over 50 million Roomba units had been sold globally~\cite{irobot2024financial}. In 2024, the market size for consumer robots increased to \$10.92 billion, with a forecast of \$40.15 billion by 2030~\cite{grandview2025consumer}. 

Like many other technologies, however, robots pose a threat to users' data privacy~\cite{Guo2022RoombaPrivacy}. Robots can autonomously navigate human environments and collect data in private spaces that users may not have originally intended or desired. Beyond traditional data privacy concerns that robots pose as a connected device collecting audio or visual data~\cite{fell2024insecure}, they introduce a new threat to users' interpersonal privacy~\cite{tang2022confidant, bell2025always}. Robots can interact with users in a human-like manner, encouraging engagement and communication. Once a robot has captured user data in this interpersonal context, it can then potentially share that data with other users. This phenomenon is particularly noteworthy in home settings where robots are placed in critical positions to engage in privacy decision making. It is within this unique context that large language models (LLMs) may hold the potential to alleviate privacy concerns. LLMs may be able to determine whether information is private, how that information should be processed, and with whom it may be shared. As privacy is highly contextual~\cite{nissenbaum2004privacy, nissenbaum2009privacy}, LLMs might help robots navigate the nuances of any given scenario.

While efforts are nascent, researchers in academia and industry are deploying LLMs and video language models (VLMs) on robots. These models equip robots with the tools to perceive and reason about their environments far beyond their previous capabilities. Researchers have demonstrated that LLMs can enhance robot capabilities by enabling natural language requests~\cite{ahn2022can}, writing code to control robot movement on-the-fly~\cite{liang2023code}, achieving interactive dialogue~\cite{shinn2023reflexion}, and improving navigation abilities~\cite{hassan2024integrating}. Several companies have also introduced projects that utilize LLMs or VLMs to enable more intelligent robots, such as Google's DeepMind-RT-2~\cite{brohan2023rt} and NVIDIA's GR00T N1~\cite{bjorck2025gr00t}. Additionally, researchers have evaluated LLM-powered robots through user studies to identify the potential benefits and weaknesses of these LLM frameworks~\cite{yuan2024towards, murali2023improving, kim2024understanding, azeem2024llm}. From these advancements, it is clear that LLMs and VLMs can offer significant benefits when integrated with robots.

Though many state-of-the-art LLMs have been designed with some level of privacy protection in mind, it is not yet clear the degree to which these models are privacy aware. Prior works have primarily focused on general LLM safety, such as LLMs' vulnerability against jailbreak attacks~\cite{zhuo2023red}. Few works, however, have investigated the \textit{privacy perceptions} of LLMs and VLMs~\cite{mireshghallah2023can, shao2024privacylens}, and none, to our knowledge, have done so in the context of human-robot interaction (HRI). Given this gap, we investigate how LLMs may help robots interpret privacy and make privacy decisions regarding data collection, processing, and sharing in home settings. Specifically, we raise the following research questions: \textbf{RQ1:} How does an individual’s privacy orientation influence their privacy expectations of robots in home settings?; \textbf{RQ2:} How well do state-of-the-art LLMs align with individuals’ privacy expectations?; and \textbf{RQ3:} How can we improve state-of-the-art LLMs’ conformity with individuals’ privacy expectations? 

Through the development and evaluation of privacy scenarios with participants and LLMs, we show that state-of-the-art LLMs maintain a broad understanding of privacy sensitivity, but may not yet possess nuanced privacy awareness. In addressing our research questions, we present the following contributions:

\begin{itemize} [leftmargin=15pt]
    \item We curate a dataset of 50 in-home scenarios grounded in contextual integrity (CI), and develop a survey to measure users' privacy preferences of robots' data collection and processing behaviors. Utilizing this dataset and survey, we conduct an online user study with 450 participants to evaluate users' preferences of robot behaviors in home scenarios.
    
    \item We assess the in-home privacy awareness of 10 state-of-the-art open-source and closed-source LLMs using four different prompting strategies. We find that while LLMs default to high privacy-preserving options, models with few-shot prompting are more capable of reflecting user preferences.
    
    \item We discuss design implications for LLM-powered robots, suggesting the potential of using LLMs as privacy controllers. By promoting data collection indicators and adopting users' privacy preferences collected through interaction, robots can enable context- and privacy-aware behavior tailored to an individual's needs.
\end{itemize}

\section{Related Works}
As we aim to explore whether user privacy preferences align with LLM-powered robots, we discuss the following threads of existing literature: (1) conceptual approaches to privacy, including privacy orientations and contextual integrity, (2) privacy concerns related to robot data collection in homes, (3) the integration of LLM reasoning in robots, and (4) LLM privacy risks and benchmarking.

\subsection{Conceptual Approaches to Privacy}
\label{sec:conceptual_approaches_to_privacy}
This work utilizes \textit{Contextual Integrity} (CI) \cite{nissenbaum2004privacy, nissenbaum2009privacy} as its primary theoretical underpinning. This theory proposes that privacy is composed of dimensions including contexts, actors, attributes, transmission principles, and purposes. These dimensions are critical for understanding the contextual dependence of what makes a situation private and how individuals perceive technology to be a threat to their privacy. Existing works have used this theory for tasks such as privacy-policy analysis \cite{sadeh2013usable, Amos_2021} and identification of privacy norms in smart homes \cite{apthorpe2018discovering}. Brause and Blank~\cite{brause2024there} have additionally proposed expanding this model to include ``information consequences'', privacy skills and awareness, and choice. In this work, we consider only the original dimensions of CI in our analysis of LLM privacy awareness.

Beyond this broad theory of privacy, many other works have attempted to assess the specific privacy attitudes, orientations, and personas of individuals as they engage with technology \cite{preibusch2013guide}. These efforts have focused on online privacy \cite{elueze2018privacy, buchanan2007development}, out-of-device privacy (\textit{i.e.}, privacy violations through direct human observation) \cite{farzand2024out}, privacy perceptions broadly \cite{chignell2003privacy, giang2023privacy}, and the identification of privacy personas \cite{hrynenko2024identifying}. Of particular note to this work is the Privacy Orientation Scale (POS) \cite{baruh2014more}, which captures an individual's perceptions of privacy as a right, concerns about their own informational privacy, other-contingent privacy, and concerns about the privacy of others. These four subscales can then be used to classify a user's privacy orientation as fitting one of three profiles, including \textit{privacy advocates}, \textit{privacy individualists}, and \textit{privacy indifferents}. Other works have described the general challenge of defining privacy personas \cite{biselli2022challenges}. Given the substantial contextual dependence of privacy, domain-agnostic assessments of privacy perceptions may also struggle to generate consistent personas across domains.

\subsection{Privacy in Robotics}
While privacy has been explored in a myriad of fields, it is a growing area of concern in robotics \cite{denning2009spotlight, guerrero2017cybersecurity, rueben2018themes, torresen2018review, pagallo2018rise, lutz2019privacy, chatzimichali2020toward}. Particular emphasis has been placed on robotics usage in homes \cite{levinson2024snitches, levinson2024our, fernandes2016detection, urquhart2019responsible}, as these environments are highly private, contain substantial amounts of sensitive information, and offer inhabitants a sense of security. Given the capacity of some robots to capture data, autonomously navigate environments, and socially engage with humans, they pose a significant threat to privacy with unique challenges compared to many other technologies. Existing works have also focused on the inherent privacy-utility trade-off that exists when robots are introduced into new environments \cite{butler2015privacy, leite2016robot}. While users may have concerns about robot privacy violations, these may be tolerated if sufficient utility is provided by the robot \cite{lutz2020robot}.

Though these concerns are widespread, many prior works have proposed solutions that attempt to limit data collection, retention, and exposure \cite{sullivan2025protecting}. For example, low-resolution sensors \cite{kim2019privacy} and purposeful sensor selection \cite{eick2020enhancing} have been proposed to limit unnecessary data collection, while sensor redirection \cite{shome2023robots} and privacy controllers \cite{tang2022confidant} have been proposed to mitigate data exposure. While these efforts have uncovered new methods for preserving human privacy, the combination of LLMs and robots presents novel opportunities for privacy-preserving LLM decision making.

\subsection{Robots \& LLM Integration}
While mainstream LLMs are a relatively recent development \cite{vaswani2017attention, radford2018improving}, LLM integration in robots has been a clear area of interest for HRI applications. Though this intersection of technologies is still in its infancy, several works have already utilized LLMs in robotics applications. These applications have been specialized, such as a chemistry assistant robot \cite{yoshikawa2023large} and a health attendant robot \cite{kim2024framework}, as well as broad, such as a `\textit{robotic brain}' to manage memory and control \cite{mai2023llm}. Other works have considered the potential of LLM-powered robots compared to other forms of LLMs (\textit{i.e.}, text- and voice-based) \cite{kim2024understanding}. All of these works, however, only begin to explore the full potential of robot-LLM integration. Further study is needed to understand the implications of LLMs for perception, control, interaction, and, most relevant to this work, decision making \cite{zeng2023large}.

\subsection{LLM Privacy Risks \& Benchmarking}
Prior works have studied LLM privacy risks and shown that the models can leak sensitive data by extracting their memory. Specifically, Carlini et al.~\cite{carlini2021extracting} have shown that a practical extraction attack could be used to recover GPT-2 training sequences. Additional works have demonstrated that LLMs can infer sensitive personal information~\cite{kim2023propile, staab2023beyond}, and remain vulnerable to prompt injection~\cite{costa2025securing, greshake2023not} and Membership Inference Attacks~\cite{duan2024membership, mattern2023membership, fu2023practical}.

To assess the potential risks and benefits that LLMs present, researchers utilize benchmarking as a standardized evaluation method for LLM performance. Common evaluation domains include reasoning~\cite{srivastava2022beyond,yi2025privacy}, multilingualism~\cite{hu2020xtreme, liang2020xglue}, coding~\cite{chen2021evaluating, austin2021program}, and instruction following~\cite{zheng2023judging, dubois2025lengthcontrolledalpacaevalsimpleway} as well as safety and privacy~\cite{ruan2023identifying, zhang2023safetybench, gu2024mllmguard, zhang2024multitrust, zharmagambetov2025agent, zhu2024privauditor, zhang2024multi, chang2025keep,bagdasarian2024airgapagent}. PURE~\cite{chi2023plue}, PolicyQA~\cite{ahmad2020policyqa}, PrivacyQA~\cite{ravichander2019question}, and GenAIPABench~\cite{hamid2023genaipabench} have been performed to evaluate privacy-policy comprehension using dedicated datasets. Mireshghallah et al.~\cite{mireshghallah2023can} first raised the question of whether LLMs can reason about CI. These authors utilized their benchmark, ConfAlde, to show that GPT-4 violates CI norms in 39\% of evaluated scenarios where humans would withhold information. Most recently, PrivacyLens~\cite{shao2024privacylens} introduced vignette-based questions to survey LLMs' privacy-norm awareness. However, existing works study whether an LLM can decide on the appropriateness of sharing data using tools from CI. In this work, we aim to explore how these models make decisions, and how well their responses align with human preferences in household HRI contexts.

\begin{table*}[b]
\caption{Dimensions of Contextual Privacy---Dimensions of privacy relevant to household HRI. These dimensions were derived from literature on Contextual Integrity and expanded to include subdimensions and factors that are relevant to HRI.}
\begin{tabularx}{\textwidth}{@{}p{0.2\textwidth} p{0.22\textwidth} X@{}}
\toprule
\textbf{Original Dimensions} & \textbf{Expanded Dimensions} & \textbf{Factors} \\
\toprule
\multirow{2}{*}{Context} 
  & Type of Location & Shared space, Private space \\
  & Robot Involvement & Robot enters part-way, Robot present from the beginning, Robot collects data from afar (\textit{i.e.}, spying and eavesdropping) \\
\midrule
\multirow{3}{*}{Actors} 
  & Number of Users & Multiple users, Single user \\
  & Age of Users & Child, Adult, Both \\
  & Relation of Users & Nuclear family member, Extended family or family friend, Outsider\\
\midrule
\multirow{2}{*}{Attributes}
  & Medium of Data & Audio of conversations, Visual of paper documents, Visual of screens, Visual of nudity (\textit{e.g.}, bathing, toileting, or intimacy) \\
  & Protected Status & Legally protected information, Non-protected information \\
\midrule
\multirow{1}{*}{Transmission Principles} 
  & Information Flow & Human to human, Human to robot, Human to no one \\
\midrule
\multirow{1}{*}{Purposes} 
  & Task Dependence & Robot maintains baseline function, Robot completes a task \\
\bottomrule
\end{tabularx}
\label{tab:privacy_dimensions}
\end{table*}

\section{Developing Privacy Scenarios}
\label{sec:dev_pri_scen}
To address our research questions, we develop a set of privacy-relevant scenarios that allow participants to report their level of concern with a robot collecting user data and their preferences for how that robot should behave following data collection. Here, we describe our process to develop these scenarios, including the use of two online user studies to crowdsource specific scenario characteristics and robot behaviors. The studies were approved by our Institutional Review Board, and all participants provided informed consent prior to their participation.

\subsection{Dimensions of Contextual Integrity}
We first reviewed existing literature on general in-home privacy concerns~\cite{choe2011living} and Contextual Integrity \cite{nissenbaum2004privacy}. From this literature, we became oriented with a comprehensive list of the many in-home activities individuals wish to keep private. Additionally, we distilled the primary dimensions of contextual privacy, including \textit{context}, \textit{actors}, \textit{attributes}, \textit{transmission principles}, and \textit{purposes}. From each of these dimensions, we further defined subdimensions relevant to privacy in HRI (\textit{e.g.}, elements that a robot may be able to identify or reason about, or that relate to its role). These expanded dimensions include \textit{type of location}, \textit{robot involvement}, \textit{number of users}, \textit{age of users}, \textit{relation of users}, \textit{medium of data}, \textit{protected status of data}, \textit{information flow}, and \textit{task dependence}. Table~\ref{tab:privacy_dimensions} presents a full list of all original dimensions, expanded dimensions, and factors.

While most of these dimensions and their factors are fairly intuitive, the factors of \textit{protected status} and \textit{medium of data} are worth discussing further. Our original intent behind protected status was to capture the sensitivity of the information involved in a given scenario. Given how subjective sensitivity may be, we chose to instead define protected status as either protected or non-protected based on legal norms. For example, information we consider to be protected includes health data~\cite{office2003summary}, academic records~\cite{ferpa}, and legal records~\cite{fre502}, among others~\cite{irc6103, glba}. Conversely, we consider information such as vacation plans or intimacy between partners to be non-protected. Additionally, we considered only specific mediums of data that we believed an LLM or VLM could reliably recognize. These data types include visuals of the human body in compromising situations (\textit{e.g.}, bathing, toileting, or intimacy), visuals of device screens, audio of conversations, and visuals of physical documents. These broad categories were selected due to the wide variety of information each of them can convey.

\subsection{Identification of Household Patterns}
With our scenario dimensions defined, we next sought to understand the frequency with which certain scenario elements would occur in a real home setting. For example, we questioned how frequently users are at home alone or with others present (\textit{i.e.}, \textit{number of users}). Given the contextual dependence of some dimensions on a robot's presence, we evaluated only the \textit{numbers of users}, \textit{types of locations}, \textit{ages of users}, and \textit{relations of users}. To achieve this goal, we conducted an online user study with 300 participants.

\subsubsection{Procedure} In this initial study, we asked participants to report the percentage of time specific qualities were true of their home (\textit{e.g.}, one person or multiple people are present in the home). We asked a series of four questions, each related to one of the dimensions described above. Each question contained a sliding scale, ranging from zero to 100 percent, for each factor within a given dimension. The sum of all sliding scales for a given question was required to equal 100 percent. Following these questions, we collected demographic information. These questions were presented in a survey developed using Qualtrics.

\subsubsection{Participants} To conduct this study, we hosted our Qualtrics survey on Prolific. Through this platform, we recruited 300 participants who were all located within the U.S. and fluent in English. Participants ranged in age from 18 to 83 years $(M = 45.88, SD = 15.95)$, and the population was split 51.33\% women, 48.00\% men, and 0.67\% non-binary. 
Participants received \$0.60 USD for completing the four-minute survey. To ensure that our data was representative of U.S. households, we additionally screened participants through Prolific to recruit a sample population consistent with the U.S. census in terms of ethnicity, sex, and age.

\subsubsection{Results} The results of this study can be found in Appendix~\ref{app:user-stats}. These values are used as weights within our scenario development do properly represent household characteristics. For all dimensions that we did not evaluate, we represent factors equally.

\subsection{Scenario Generation}
\begin{figure}
    \centering
    \includegraphics[width=\linewidth]{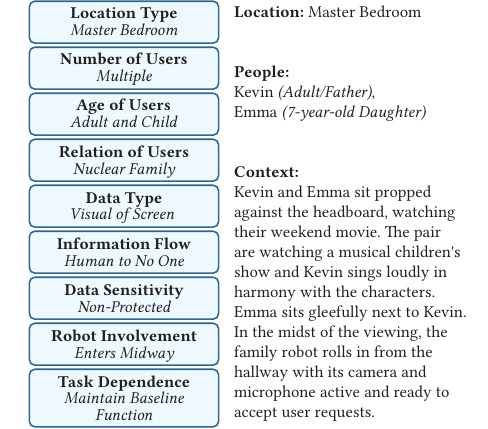}
    \caption{\textit{Scenario Script ---} An example script showcasing a robot capturing a private moment within the home. Each script is developed based on our nine dimensions of privacy grounded in Contextual Integrity.}
    \label{fig:script}
\end{figure}

After defining our scenario dimensions and crowdsourcing frequency weights, we next generated our privacy scenarios. Similar to the scenario scripts developed by Tang et al.~\cite{tang2022confidant}, we formatted ours to include location, characters, contextual information to describe the scene, and, in most cases, a scripted dialogue between characters. Using the dimensions defined above, we first created a set of all possible combinations of dimensions and then randomly sampled 50 combinations consistent with our crowdsourced weights. Given our sample of dimension combinations, we prompted ChatGPT 4o, Gemini 2.0 Pro, and Claude Sonnet 3.7 to generate the initial iterations of our privacy-relevant scenarios. By generating the scenarios, we hoped to represent a more diverse set of scenario characteristics than we could develop on our own. We next verified that each scenario properly represented the dimensions upon which it was based and made corrections when needed. We additionally edited the scenarios to remove any explicit references to privacy, such that neither the participants nor the studied LLMs would be primed. We present an example script in Figure~\ref{fig:script}. The full dataset, survey, and the prompt used to generate the scenarios, can be accessed in our OSF repository\footnote{\url{https://osf.io/wbp37/?view_only=4a2799ee7cf04480a54d06faf342714d}}.

\subsection{Crowdsourcing Robot Behaviors}
For each of our developed scenarios, we hoped to capture participant preferences for the robot's behavior in response to its collection of user data. We did not, however, have a set of standardized robot response options that we could present to participants and, later, our selected LLMs. Therefore, we conducted a second online study to crowdsource these options for robot responses. While we could have developed these behaviors on our own, we felt that participant responses would better reflect a diverse set of user-desired behaviors and avoid author bias.

\subsubsection{Procedure} In this survey, each participant was assigned one of two conditions. In the first condition, participants were told to imagine themselves going about their daily routines at home when their robot records private information about them. In the second condition, participants were told to imagine the same scenario, however, they are provided additional hypothetical detail to ground their responses (\textit{i.e.}, \textit{``This scenario may involve you alone, your interactions with other people, or interactions with the robot, and may occur in shared or private spaces within the home.''}). We created these two conditions to collect responses that were either entirely unbiased or grounded in context. Participants were then asked how they would prefer the robot to behave internally (\textit{i.e.}, how it would process and manage collected data) and outwardly (\textit{i.e.}, how it would interact with the participant). Following these questions, we collected demographic information.

\subsubsection{Participants} We hosted this study on Prolific and recruited 62 participants (31 per condition). All participants were located within the U.S. and fluent in English. Participant ages ranged from 20 to 74 $(M = 45.42, SD = 14.95)$ and the population was split 48\% women, 50\% men, and 2\% non-binary. Participants received \$1.00 USD for completing the six-minute survey. Rather than collecting data from a predefined number of participants, we collected data until our responses reached saturation (\textit{i.e.}, no new behavior types were identified after several consecutive participants)~\cite{hennink2022sample}.

\subsubsection{Analysis} After collecting our study data, the first author reviewed each response and developed an initial codebook that contained individual codes that were grouped within overarching thematic categories~\cite{clarke2014thematic}. Then, the third author independently coded the responses again using the codebook. Afterwards, both authors discussed the coded responses and made modifications to the codes and categories until consensus was reached. The final categories of codes are as follows: data collection, user notification, data management, and data sharing. These categories and codes were then used to develop questions that were ultimately used in participant and LLM evaluations of our privacy scenarios. The full codebook can be found in Appendix~\ref{app:codebook}.

\section{User Evaluation}
\label{sec:user_study}
In this section, we address \textbf{RQ1} through a large online user study to evaluate which robot behaviors are preferred by users within a household HRI context. This study utilizes the privacy scenarios and robot behaviors defined above as a standardized set of stimuli to evaluate with both users and, as we will see in Section~\ref{sec:llm_eval}, LLMs.

\subsection{Procedure}
From our scenarios and robot behaviors, we developed a survey on Qualtrics that we hosted on Prolific. This survey began with a simple overview of the enclosed questions, instructions on how to complete them, and a consent form. Participants were informed that by continuing past the consent page, they were consenting to take part in the study. Next, participants were asked to complete the Privacy Orientation Scale (POS) \cite{baruh2014more}. Following this scale, participants were randomly assigned three of our 50 privacy scenarios. After reading each scenario, participants responded to a series of 15 questions related to the privacy sensitivity of the scenario and the behaviors they preferred the robot to utilize in response to that scenario. Participants were prompted to answer these questions from the perspective of the characters involved in the scenarios. Finally, participants completed a series of demographic questions.

\subsection{Measures}
This study utilizes the POS \cite{baruh2014more} as well as a set of 15 items developed by the research team focusing on preferred robot responses. The POS is composed of 16 items rated on a five-point Likert scale, which aggregate to four subscales (\textit{i.e.}, Subscale 1: Privacy as a right, Subscale 2: Concern about own informational privacy, Subscale 3: Other-contingent privacy, and Subscale 4: Concern about privacy of others). These subscales allow us to understand participants' privacy predispositions as they relate to preferred robot behaviors.

The additional 15 items we included for each scenario are based on our preliminary study on preferred robot responses (see Section \ref{sec:dev_pri_scen}). We first adapted our robot-response codes and added additional robot responses that were not provided by participants during our online study (\textit{e.g.}, redirecting sensors or leaving a room when sensitive data is identified). We also removed codes that were not relevant (\textit{e.g.}, “Discomfort with robot” was a common sentiment, but not a valid robot behavior). We next utilized these responses as the answers for a set of binary and categorical questions. The binary items included both \textit{privacy-enhancing} and \textit{non-interfering} or \textit{non-restrictive} robot responses (\textit{e.g.}, \textit{tell you that data has been collected} or \textit{do not tell you that data has been collected}). All categorical items included an ordinal spectrum of responses (\textit{e.g.}, \textit{delete data immediately}, \textit{save data only temporarily and delete after a predefined period}, and \textit{save data indefinitely}). One additional question asked participants to rate their privacy concerns with a given scenario on a scale from one to ten. We further performed a regression analysis on how participants' POS scores relate to their 15 additional scenario-based questions: one about scenario sensitivity and 14 about preferred robot responses (referred to as outcome variables in Figure \ref{fig: regression_analysis}).

\subsection{Participants}
We recruited 450 participants through Prolific, each of whom evaluated three of the 50 scenarios (27 participants per scenario). This balance was chosen to ensure generalizability across a breadth of scenarios while maintaining sufficient statistical power~\cite{westfall2014statistical}. All participants were located within the U.S. and fluent in English. Participant ages ranged from 19 to 88 years $(M = 40.69, SD = 13.43)$ and the population was split 46.67\% women, 52.67\% men, 0.44\% non-binary, and 0.22\% prefer not to say. Participants were compensated \$4.00 USD for completing the 16-minute survey.

\begin{figure}
    \centering
    \includegraphics[width=\linewidth]{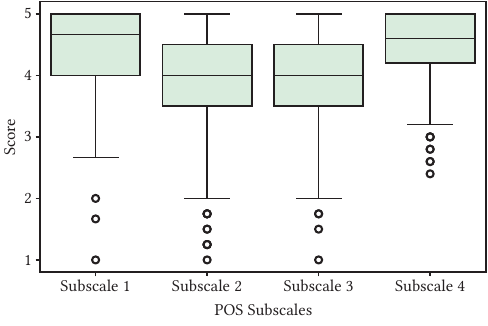}
    \caption{\textit{Participant POS Scores---}Box plots presenting the POS scores of all 450 participants across each subscale (\textit{i.e.}, Subscale 1: Privacy as a right, Subscale 2: Concern about own informational privacy, Subscale 3: Other-contingent privacy, and Subscale 4: Concern about privacy of others).}
    \label{fig:participant_pos}
\end{figure}

\subsection{Results}
Next, we present the results of our user evaluation, including participant privacy orientations, preferred robot responses to scenarios, and a regression analysis examining the degree to which the former predicts the latter.

\subsubsection{Privacy Orientation}
\label{sec:privacy_orientation}
After calculating POS scores for each of our 450 participants, we see that participants are highly privacy oriented. On average, participants scored 4.49 (SD = 0.61) on Subscale 1, 3.84 (SD = 0.81) on Subscale 2, 3.96 (SD = 0.77) on Subscale 3, and 4.50 (SD = 0.52) on Subscale 4 (see Figure \ref{fig:participant_pos}). These scores are fairly consistent with the original POS scores reported by Baruh and Cemalc{\i}lar~\cite{baruh2014more}, though they are all consistently elevated. We additionally see lower, but acceptable, reliability among our sample compared to that of the original POS validation. A full listing of the mean, standard deviation, and reliability of each subscale in our sample and that of Baruh and Cemalc{\i}lar~\cite{baruh2014more} is in Appendix \ref{app:POS_stats}.

We additionally conducted a two-step cluster analysis using Schwarz’s Bayesian Information Criterion (BIC) to determine whether the same three clusters would emerge as shown by Baruh and Cemalc{\i}lar~\cite{baruh2014more}. While our analysis is consistent with that of Baruh and Cemalcılar (\textit{i.e.}, good fit and a 0.5 silhouette measure of cohesion and separation), only two clusters emerged. One cluster is consistent with the \textit{privacy advocates} cluster (\textit{i.e.}, highest scores for subscales one and four among identified clusters) and the other is consistent with the \textit{privacy indifferents} cluster (\textit{i.e.}, low scores on all four subscales). These clusters, however, are highly skewed with 96\% of participants fitting within the \textit{privacy advocates} cluster and 4\% in the \textit{privacy indifferents} cluster. While it is unclear why ours differ, prior work has previously described the difficulty of generating privacy profiles \cite{biselli2022challenges}. Given the contextual dependence of privacy attitudes and preferences, a single privacy profile may not be able to capture the broad spectrum of beliefs an individual has across specific domains. Given the low variability of profile membership in our sample (\textit{i.e.}, nearly all participants were in the same cluster), our subsequent analyses examine participants' raw scores on each of the four subscales rather than participants' global profile membership.

\begin{figure}[t]
    \centering
    \includegraphics[width=\linewidth]{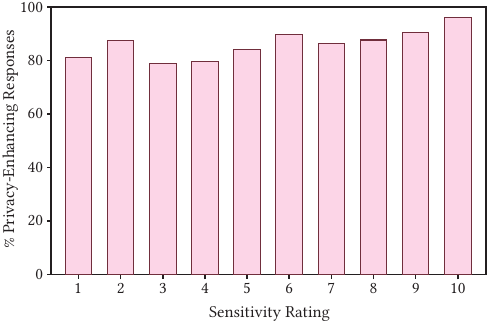}
    \caption{\textit{Privacy-Enhancing Preferences---}Aggregate scores of all binary response items (\textit{e.g.}, privacy-enhancing or non-interfering) across scenario sensitivity ratings.}
    \label{fig:privacy_enhancing_percentages}
\end{figure}

\subsubsection{Preferred Robot Responses}
Throughout the course of our study, each of our 450 participants evaluated three privacy scenarios, for a total of 1350 evaluations. Given the scale and complexity of this data, here we present broad descriptive statistics about the dataset. When we control for perceived scenario sensitivity, we see that participants most commonly prefer privacy-enhancing robot responses (\textit{i.e.}, in terms of notification behaviors, processing and storage of data, and discussion of data management) across all scenario sensitivity levels. We additionally see that as perceived scenario sensitivity increases, a greater percentage of participants prefer privacy-enhancing behaviors (see Figure \ref{fig:privacy_enhancing_percentages}). These data show a clear preference for privacy-preserving robot responses regardless of scenario or data sensitivity.

Regarding all remaining response behaviors (\textit{i.e.}, avoidance of further data collection, immediate retention decisions, retention duration, and sharing policies), we again see fairly consistent responses regardless of scenario sensitivity. When considering avoidance of data collection, participants most frequently preferred the robot to deactivate its sensors at all sensitivity levels except the lowest. In this case, participants preferred that the robot make no change to its behavior. These results suggest that participants have a low tolerance for privacy violations of all but the very least sensitive kind. Regarding immediate retention decisions and retention duration, participants preferred the robot to retain only essential data and to delete that data after a predefined period, respectively, for nearly all sensitivity levels. At sensitivity levels of nine and ten, participants preferred that the robot not retain any information (immediate retention). At sensitivity levels of eight and greater, participants preferred that data be deleted immediately (retention duration). These results imply that participants are amenable to the possibility that robots may require certain types of data to maintain functionality. Regarding sharing policies, participants consistently preferred that data only be shared with users to whom the data is directly relevant, regardless of scenario sensitivity. Given that this option was the most stringent available response, this result again suggests a low tolerance for even minor privacy violations.

\subsubsection{Regression Analysis}
\label{sec:regression_analysis}
\begin{figure}
    \centering
    \includegraphics[width=\columnwidth]{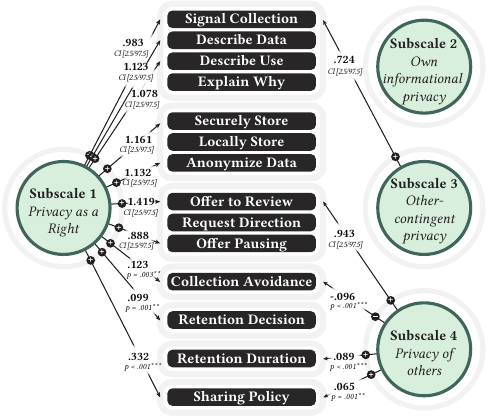}
    \caption{\textit{Regression Analysis---}Relationships between the four POS subscales and 14 outcome variables (\textit{i.e.}, participants' preferred robot responses to privacy scenarios).}
    \label{fig: regression_analysis}
\end{figure}

To understand the relationship between participants' POS scores and preferred behaviors (\textit{i.e.}, \textbf{RQ1}), we created regression models utilizing each of the four POS subscales as predictors and each type of response behavior as outcome variables. We utilized generalized linear mixed models for binary response questions and linear mixed models with Bonferroni correction for categorical response questions. All of the linear mixed models used ordinal outcome variables that were converted to integer values. While some other models are able to utilize categorical data by default, we felt a linear mixed model would better account for the mixed effects inherent to our dataset and remain interpretable. In total, we developed 14 models, one for each type of response behavior. Here we present the results of our regression analyses. A visualization of all 14 models can be seen in Figure \ref{fig: regression_analysis}.

\textit{Subscale 1: Privacy as a Right.}
Across nearly all of our regression models, we can see that Subscale 1, the perception of \textit{privacy as a right}, is a strong predictor. Subscale 1 predicts three notification behaviors (\textit{i.e.}, signaling active data collection, informing a user of the data collected, and describing what will happen to that data), two storage or processing behaviors (\textit{i.e.}, securely storing data and anonymizing data), two management discussion behaviors (\textit{i.e.}, offering a user to review and manage data and offering to pause data collection), as well as behaviors related to avoidance of further data collection, retention or purging of data, retention durations, and sharing policies. It is clear that the perception that privacy is a right is a strong indicator of whether a person prefers these privacy-preserving behaviors. While it is unclear why Subscale 1 did not predict other behaviors, such as storing data locally, it is possible that participants do not fully understand the implications of local storage compared to cloud storage. Other behaviors, such as asking a user how to manage data, may not be seen as essential by comparison to notifying a user that data has been collected or understanding with whom that data may be shared.

\textit{Subscale 2: Concern for Personal Informational Privacy.}
By contrast to Subscale 1, Subscale 2, or \textit{concerns about own informational privacy} did not predict any of the outcome variables we analyzed. This may be because participants were asked to answer questions from the perspective of the characters in our scenarios rather than from their own perspectives. This result may reflect a disconnect between participants' reported POS scores and the behaviors they would prefer were they the characters in the analyzed scenes.

\textit{Subscale 3: Other-Contingent Privacy.}
The next subscale, \textit{other-contingent privacy}, predicts only a single outcome variable: explaining why data was collected. Considering this subscale refers to the ways others regard privacy (\textit{i.e.}, how others' care for privacy may impact one's own), it stands to reason that a person concerned with this concept would prefer to understand the reasoning behind another agent's data collection. While other outcome variables, such as sharing policy, may also inform a user's understanding of the robot's privacy perspective, an explanation as to why data was collected may be far more informative to users.

\textit{Subscale 4: Concern for the Privacy of Others.}
Our final predictor is Subscale 4, or \textit{concern about privacy of others}. This subscale predicts four outcome variables: offering users the ability to review and manage data, avoiding further data collection, retention duration, and sharing policy. Compared to many of the other outcome variables, these appear to have the greatest impact on the direct harm that a robot can potentially inflict on a user's privacy. For instance, the robot may continue collecting data, retain it for longer than is necessary, or expose that data to others. Offering the user the ability to review and manage that data returns control to that user. In this way, these outcomes may be considered the most critical to protecting the privacy of those involved in the analyzed scenarios. It is not clear why an outcome like retention decision (\textit{deciding whether to immediately delete or retain data}) is not also predicted by Subscale 4, but further exploration may clarify this relationship.

\section{LLM Evaluation}
\label{sec:llm_eval}
To address our remaining research questions (\textit{i.e.}, \textbf{RQ2} and \textbf{RQ3}), we examine (1) how LLMs respond to our developed scenarios and questions, (2) how their responses change when we utilize various prompting strategies, and (3) how the best-performing model reasons when responding. Our prompting strategies include providing only scenarios and questions for the LLMs to respond to (\textit{i.e.}, default or \textit{zero-shot}), providing participant POS scores (\textit{i.e.}, POS prompting), providing participant ratings of sensitivity (\textit{i.e.}, sensitivity prompting), providing participant POS scores and sensitivity ratings (\textit{i.e.}, POS \& sensitivity prompting), and providing POS scores, sensitivity ratings and an example scenario (\textit{i.e.}, few-shot promoting). Here, we present the details of the models we examine, our prompting procedures, and the results of our subsequent analysis.

\begin{figure*}
    \centering
    \includegraphics[width=\linewidth]{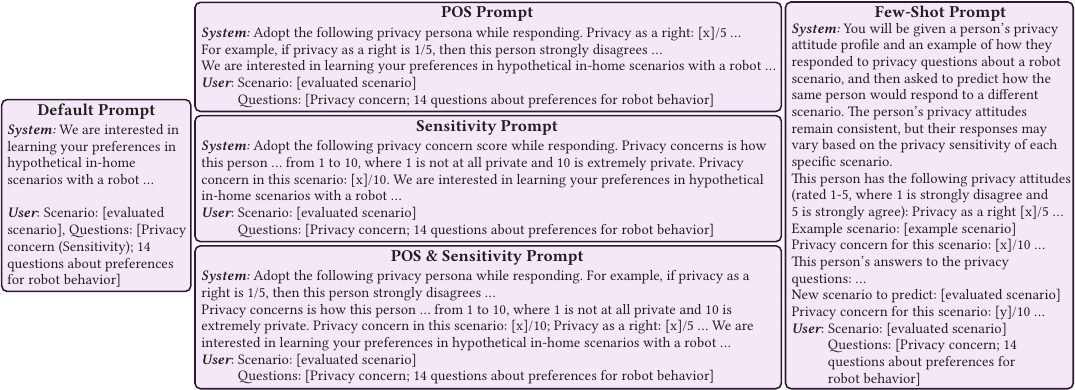}
    \caption{\textit{Examples of Prompting Strategies---}Prompt examples for default, POS, sensitivity, POS and sensitivity, and few-shot prompting. These strategies each include unique system prompts, followed by the same user prompt.}
    \label{fig:prompts}
\end{figure*}

\subsection{Model Selection \& Setup}
We evaluated five leading closed-source models, including GPT-o3 (o3-2025-04-16)~\cite{openai_o3_2025}, GPT-4o (gpt-4o-2024-08-06)~\cite{openai_gpt4o_2025}, Gemini 2.5-Flash (gemini-2.5-flash-preview-04-17)~\cite{google_gemini_2.5_flash_2025}, Gemini 2.5-Pro (gemini-2.5-pro-preview-05-06)~\cite{google_gemini_2.5_pro_2025}, and Grok 3 (grok-3-beta)~\cite{xai_grok_models_2025} as well as five open-sourced models, including DeepSeek V3 (deepseek-chat)~\cite{deepseek_api_pricing_2025}, Qwen 3-32B~\cite{yang2025qwen3}, Mistral 7B-Instruct-v0.3~\cite{mistral2024mistral7b}, Llama 3.3-70B-Instruct~\cite{meta2024llama33}, Llama 4 Scout-17B-16E-Instruct~\cite{meta_llama4_scout_2025}.

As we assess each model, we utilize similar parameters; we set the temperature to 0.3 and the reasoning effort to `low' for reasoning models. For each selected model, we run three trials to verify consistency in responses. We then accept the majority response for each item between the three trials. Two of the open-source models (\textit{i.e.}, Qwen and Mistral) are evaluated over two A6000 48GB GPUs, and the remaining models are evaluated through their respective APIs (Huggingface Inference API and kluster.ai for the Llama series). For any given model or trial, we prompt the LLM to respond in a JSON format.

\subsection{Default Prompting Conformance}
In this phase, we explore the privacy-related decision making of out-of-the-box LLMs and the degree to which their responses conform to user expectations. To begin, we first compare each LLM's responses with those of our participants. This process sets a baseline expectation for subsequent evaluation during which we provide the LLMs with elements of the users' privacy attitudes and preferences. To properly compare the LLM responses with those of our participants, we present instructions to the LLMs that strictly resemble those provided during our user study. To that end, our model inputs are composed of a system prompt, containing any necessary background information and output format, as well as a user prompt, containing both scenarios and questions. A detailed example prompt is presented in Figure~\ref{fig:prompts}. Following data collection, we are left with responses to 15 survey questions for each of our 50 scenarios across each of the evaluated LLMs.

\subsubsection{Analysis Procedure}
To compare the data from a single LLM (\textit{i.e.}, one response per each of 50 scenarios) to the data of all participants (\textit{i.e.}, 27 responses per each of 50 scenarios), we expanded the LLM data to mirror our full participant data set. We duplicate the LLM data 27 times for ease of comparison across the 27 participants per scenario. Next, we compute Micro and Macro F1 scores between the participant and LLM data, treating the human responses as the ground truth. To compare numerical data (\textit{i.e.}, sensitivity rating), we calculate per-scenario mean absolute error (MAE).

\begin{table}[t]
\caption{Default LLM Performance---Agreement between LLM and participant response, separated by binary and categorical items, in terms of Micro and Macro F1 scores.}
    \centering
    \begin{tabular}{lcccc}
        \toprule
        \multirow{2}{*}{\textbf{Model}} & \multicolumn{2}{c}{\textbf{Binary F1}} & \multicolumn{2}{c}{\textbf{Categorical F1}} \\
        \cmidrule(lr){2-3} \cmidrule(lr){4-5}
        & Macro & Micro & Macro & Micro\\
        \midrule
        Llama 4 Scout & 0.464 & \textbf{0.868} & \textbf{0.286} & \textbf{0.531}  \\
        GPT-o3 & 0.464 & \textbf{0.868} & 0.266 & 0.514\\
        Gemini 2.5-Pro & \textbf{0.472} & 0.863  & 0.257 & 0.489  \\
        GPT-4o & 0.464 & \textbf{0.868} & 0.230 & 0.493\\
        DeepSeek V3 & 0.464 & \textbf{0.868} & 0.238 & 0.481  \\
        Gemini 2.5-Flash & 0.466 & 0.866  & 0.236 & 0.483 \\
        Grok 3 & 0.464 & \textbf{0.868} & 0.230  & 0.474  \\
        Llama 3.3-70B & 0.464 & \textbf{0.868} & 0.213 & 0.484  \\
        Qwen 3-32B & 0.467 & 0.865 & 0.202  & 0.474  \\
        Mistral 7B-Instruct & 0.414 & 0.711  & 0.137 & 0.357  \\
        \bottomrule
    \end{tabular}
    \label{tab:default_f1_binary_categorical_multi}
\end{table}

\subsubsection{Results}
We present our results in Tables \ref{tab:default_f1_binary_categorical_multi} and \ref{tab:mae_default_subscale}. For comparison, we separate these results into three categories: binary (\textit{e.g.}, privacy-enhancing or non-interfering), categorical (\textit{i.e.}, three or more response options), and numerical (\textit{i.e.}, scaled from 1 to 10). Overall, performance between models is relatively consistent, except for Mistral, which performs notably worse among binary and categorical items. Across models, high Micro F1 scores indicate that the models perform fairly well when compared to common user responses (\textit{i.e.}, when participants prefer privacy-enhancing behaviors). Low Macro F1 scores, however, indicate that the models perform worse on rare responses (\textit{i.e.}, when participants prefer non-intrusive behaviors). While this pattern is present among all data, weaknesses in model performance become more pronounced in categorical data due to the increased number of possible responses. We also observe that six models show a perfect tie among binary items, as these models selected privacy-enhancing choices for all items. 

\begin{figure}[b]
    \centering
    \includegraphics[width=\linewidth]{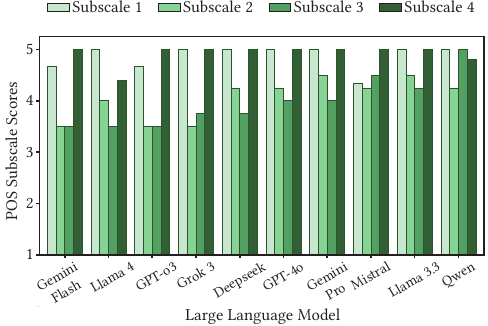}
    \caption{\textit{LLM POS Scores---}Privacy orientation scores for each of the evaluated LLMs. Among subscales related to user privacy (\textit{i.e.}, 1 and 4), nearly all LLMs produce a maximal score.}
    \label{fig:llm_pos}
\end{figure}

To further investigate why models prefer privacy-enhancing responses by default, we additionally examine the privacy orientation of the LLMs. As shown in Section~\ref{sec:user_study}, participant privacy orientation influences response preferences. Therefore, we queried models to respond to the POS \cite{baruh2014more} in the same manner we presented this scale to participants. The results of this process are presented in Figure~\ref{fig:llm_pos}. We see that nearly all models rate \textit{privacy as a right} at the maximum (\textit{i.e.}, 5.00), except Gemini 2.5 Flash, GPT-o3, and Mistral (4.67, 4.67, and 4.33, respectively). Similarly, nearly all models rate \textit{concern about privacy of others} at the maximum, except Qwen and Llama 4 Scout (4.80 and 4.40, respectively). Among the remaining two subscales, \textit{concern about own informational privacy} and \textit{other-contingent privacy}, which are less meaningful from the perspective of an LLM, all models scored between 3.50-4.50 and 3.50-5.00, respectively. To the extent that these results can be interpreted as privacy orientations of the LLMs, they indicate that all models are highly concerned with user privacy.

\begin{table}
\caption{MAE of LLM Sensitivity Ratings---Mean Absolute Error (MAE) of LLM sensitivity ratings compared to participant ratings under default and POS prompting. These ratings show that the addition of participant POS may both help or hinder LLM sensitivity conformity.}
    \centering
    \begin{tabular}{lccc}
        \toprule
        \multirow{2}{*}{Model} 
          & \multicolumn{2}{c}{MAE $\downarrow$} 
          & \multirow{2}{*}{\begin{tabular}{@{}c@{}}Percent\\Reduction (\%)\end{tabular}} \\
        \cmidrule(lr){2-3}
          & Default & POS &  \\ 
        \midrule
        Qwen 3-32B        & 3.266 & \textbf{2.573} & \phantom{$+$}21.2 \\
        DeepSeek V3       & 3.178 & 2.822 & \phantom{$+$}11.2 \\
        Grok 3            & 2.950 & 2.884 & \phantom{$+$}2.2  \\
        Gemini 2.5-Flash  & 3.370 & 3.376 & \phantom{$-$}--0.2 \\
        Llama 3.3-70B     & 2.767 & 2.801 & \phantom{$-$}--1.2 \\
        Llama 4 Scout     & \textbf{2.661} & 2.778 & \phantom{$-$}--4.4 \\
        GPT-4o                & 2.773 & 3.046 & \phantom{$-$}--9.9 \\
        Gemini 2.5-Pro    & 3.102 & 3.470 & \phantom{$-$}--11.8 \\
        Mistral 7B-Instruct  & 2.669 & 2.984 & \phantom{$-$}--11.8 \\
        GPT-o3                & 2.806 & 3.244 & \phantom{$-$}--15.6 \\
        \bottomrule
    \end{tabular}
    \label{tab:mae_default_subscale}
\end{table}

\subsection{Conformity Under Prompting Strategies}
We next explore prompting strategies that have the potential to improve model conformity to human preferences. All strategies involve providing the LLMs with information about the participants and subsequently requesting responses to scenarios based on the participants' perspectives. This information includes participant POS scores, participant ratings of scenario sensitivity, as well as POS scores \textit{and} sensitivity ratings. We additionally utilize few-shot prompting as our final strategy. In our few-shot prompts, we provide the LLMs with an example participant response (\textit{i.e.}, participant POS, sensitivity rating, and responses regarding preferred robot behaviors for a single scenario). Examples of these prompting strategies can be seen in Figure \ref{fig:prompts}. We present the resulting performance of each of these models and strategies in Table~\ref{tab:prompt_f1_binary_categorical_multi}. Comparisons between prompting strategies can be seen in Table \ref{tab:t-test}.

\subsubsection{POS Prompting}
The first prompting strategy we use involves providing the LLMs with participant POS scores. As we have shown in Section \ref{sec:regression_analysis}, these scores, particularly Subscale 1, are predictive of various preferences for robot responses. We therefore anticipate that the LLMs will be more capable of aligning their responses to participants with this information. Consequently, we generate responses that utilize the POS scores of each participant across the specific scenarios each participant evaluated. As a result, we obtain a dataset for each LLM that mirrors the dataset provided by participants. We again compute Micro and Macro F1 scores, separated by binary and categorical data, for each LLM.

\begin{figure}[ht]
    \centering
    \includegraphics[width=\linewidth]{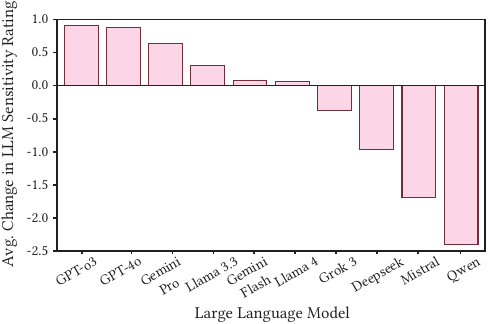}
    \caption{\textit{Shift in LLM Sensitivity Ratings---}Compared to default prompting, some LLMs increase sensitivity ratings and others decrease these ratings to varying degrees following POS prompting.}
    \label{fig:sensitivity_shift}
\end{figure}

\textit{Results.}
Compared to default model performance, we primarily see modest and non-significant changes in conformity under this prompting strategy (see Table~\ref{tab:t-test}). As part of our analysis, we also examine how this prompting strategy influences LLM perceptions of scenario sensitivity. We do not observe a substantial improvement in this regard among the evaluated models, except for Qwen (see Table~\ref{tab:mae_default_subscale}). Moreover, several models even exhibit an increase in mean absolute error (MAE). The models that exhibit the smallest shift (see Figure~\ref{fig:sensitivity_shift}), whether positive or negative, are the Llama series, the Gemini 2.5 Flash, and Grok3. These results suggest that even if user POS scores are provided to a model, they may mislead the model in terms of assessing the sensitivity of a given scenario. This possibility motivates our next prompting strategy, sensitivity prompting.

\subsubsection{Sensitivity Prompting}
As participant and LLM sensitivity ratings vary widely, even when providing the LLMs with participant POS scores, we next provide the LLMs with additional participant sensitivity ratings directly for each participant-scenario pair.

\textit{Results.}
We see a statistically significant increase in Macro F1 scores and a statistically significant decrease in Micro F1 scores for both binary and categorical data (see Table~\ref{tab:t-test}). These results align with patterns we observed in the default prompting results. In the vast majority of cases, LLMs choose privacy-preserving responses by default and subsequently perform well on common cases (\textit{i.e.}, participants choose privacy-enhancing responses) but poorly on rare cases (\textit{i.e.}, participants choose non-privacy-enhancing responses). This outcome is reflected in the consistently higher Micro F1 scores over Macro F1 scores. When the LLMs are then provided participant sensitivity ratings, Micro F1 scores decrease, while Macro F1 scores increase. This outcome suggests that the LLMs begin to see the scenarios as less private and respond accordingly.

\begin{table*}
\caption{LLM Performance by Prompting Strategy---Conformity of LLMs to participant responses based on our four prompting strategies (\textit{i.e.}, POS prompting, sensitivity prompting, POS \& sensitivity prompting, and few-shot prompting). These results include Micro and Macro F1 scores and are separated based on binary and categorical items.}
    \centering
    \scriptsize
    \setlength{\tabcolsep}{4pt}
    \begin{tabular}{@{}>{\raggedright\arraybackslash}p{1.7cm}*{16}{c}@{}}
        \toprule
        \multirow{3}{*}{\textbf{Model}}
            & \multicolumn{4}{c}{\textbf{POS only}}
            & \multicolumn{4}{c}{\textbf{Sensitivity only}}
            & \multicolumn{4}{c}{\textbf{POS + Sensitivity}}
            & \multicolumn{4}{c}{\textbf{Few-Shot}} \\
        \cmidrule(lr){2-5}\cmidrule(lr){6-9}\cmidrule(lr){10-13}\cmidrule(lr){14-17}
            & \multicolumn{2}{c}{Binary F1}
            & \multicolumn{2}{c}{Categorical F1}
            & \multicolumn{2}{c}{Binary F1}
            & \multicolumn{2}{c}{Categorical F1}
            & \multicolumn{2}{c}{Binary F1}
            & \multicolumn{2}{c}{Categorical F1}
            & \multicolumn{2}{c}{Binary F1}
            & \multicolumn{2}{c}{Categorical F1} \\
        \cmidrule(lr){2-3}\cmidrule(lr){4-5}
        \cmidrule(lr){6-7}\cmidrule(lr){8-9}
        \cmidrule(lr){10-11}\cmidrule(lr){12-13}
        \cmidrule(lr){14-15}\cmidrule(lr){16-17}
            & Macro & Micro & Macro & Micro
            & Macro & Micro & Macro & Micro
            & Macro & Micro & Macro & Micro
            & Macro & Micro & Macro & Micro \\
        \midrule
        Grok 3                  & 0.483 & 0.854 & 0.278 & 0.507 & 0.513 & 0.650 & 0.319 & 0.431 & 0.517 & 0.826 & 0.332 & 0.515 & 0.707 & \textbf{0.885} & 0.475 & 0.626 \\
        GPT-o3                      & 0.470 & 0.862 & 0.268 & 0.503 & 0.508 & 0.640 & 0.326 & 0.419  & 0.525 & 0.790 & \textbf{0.358} & \textbf{0.523} & \textbf{0.716} & 0.879 & 0.454 & 0.586 \\
        Llama 3.3-70B           & 0.497 & 0.852 & 0.274 & \textbf{0.514} & 0.511 & 0.665 & \textbf{0.328} & 0.465 & 0.533 & 0.770 & 0.304 & 0.497 & 0.661 & 0.823 & 0.461 & 0.593 \\
        GPT-4o                      & 0.491 & 0.859 & 0.248 & 0.503 & 0.513 & 0.641 & 0.314 & 0.440  & 0.512 & 0.834 & 0.263 & 0.486 & 0.650 & 0.807 & \textbf{0.486} & \textbf{0.641} \\
        DeepSeek V3             & \textbf{0.501} & 0.850 & 0.261 & 0.467 & 0.518 & 0.672 & 0.274 & 0.396 & 0.527 & 0.817 & 0.286 & 0.461 & 0.683 & 0.839 & 0.475 & 0.610 \\
        Llama 4 Scout            & 0.484 & 0.858 & \textbf{0.279} & 0.506 & 0.510 & 0.642 & 0.283 & \textbf{0.492} & 0.524 & 0.759 & 0.310 & 0.511 & 0.595 & 0.745 & 0.454 & 0.589 \\
        Gemini 2.5-Pro          & 0.479 & 0.864 & 0.225 & 0.473 & 0.502 & 0.616 & 0.307 & 0.412  & 0.532 & 0.777 & 0.310 & 0.470 & 0.677 & 0.859 & 0.440 & 0.586 \\
        Qwen 3-32B              & 0.465 & \textbf{0.868} & 0.184 & 0.448 & \textbf{0.534} & 0.812 & 0.245 & 0.442 & 0.490 & \textbf{0.842} & 0.206 & 0.447 & 0.662 & 0.838 & 0.432 & 0.577 \\
        Gemini2.5-Flash        & 0.496 & 0.852 & 0.208 & 0.463 & 0.512 & 0.648 & 0.323 & 0.423  & \textbf{0.534} & 0.721 & 0.285 & 0.445 & 0.681 & 0.846 & 0.455 & 0.584 \\
        Mistral 7B-Instr & 0.477 & 0.821 & 0.235 & 0.339  & 0.485 & \textbf{0.851} & 0.170 & 0.447 & 0.460 & 0.764 & 0.172 & 0.377 & 0.567 & 0.865 & 0.421 & 0.582 \\
        \bottomrule
    \end{tabular}
    \label{tab:prompt_f1_binary_categorical_multi}
\end{table*}

\subsubsection{POS \& Sensitivity Prompting}
To further improve model performance, we provide the LLMs with both participants' POS scores and sensitivity ratings. Previously, POS prompting did not yield improved model performance and, in many cases, worsened model sensitivity ratings. Through sensitivity prompting, we saw increases in Macro F1 scores and decreases in Micro F1 scores. By providing both participant POS scores and sensitivity ratings, we anticipate that the LLMs will be able to utilize POS scores more effectively without hindering perceptions of scenario sensitivity.

\begin{table}[b]
\caption{Comparative Prompting Strategy Performance---Aggregate differences and t-tests of model performance by default and while using prompting strategies.}
\centering
\setlength{\tabcolsep}{3.5pt}
\scriptsize
\begin{tabular}{llcccc}
\toprule
\multirow{2}{*}{\textbf{Prompting}} & \multirow{2}{*}{\textbf{Statistic}} 
& \multicolumn{2}{c}{\textbf{Binary}} 
& \multicolumn{2}{c}{\textbf{Categorical}} \\
\cmidrule(lr){3-4} \cmidrule(lr){5-6}
\textbf{Strategy} & & \textbf{Macro F1} & \textbf{Micro F1} & \textbf{Macro F1} & \textbf{Micro F1} \\
\midrule
\multirow{0.2}{*}{POS}
& Difference & 2.38\% & 0.27\%  & 1.66\%  & -0.55\% \\
& t-test     
  & \textbf{\boldmath$p=.003$} & $p=.826$   
  & $p=.242$ & $p=.449$   \\
\midrule
\multirow{0.2}{*}{Sensitivity} 
& Difference & 5.00\% & -16.74\% & 5.93\%  & -4.11\% \\
& t-test     
  & \textbf{\boldmath$p=.000$} & \textbf{\boldmath$p=.002$}   
  & \textbf{\boldmath$p=.000$} & \textbf{\boldmath$p=.034$} \\
\midrule
\multirow{0.2}{*}{POS \& Sens.} 
& Difference & 5.49\% & -6.12\%  & 5.31\%  & -0.45\% \\
& t-test     
  &\textbf{\boldmath$p=.000$} & \textbf{\boldmath$p=.007$}   
  & \textbf{\boldmath$p=.001$} & $p=.577$   \\
\midrule
\multirow{0.2}{*}{Few-Shot} 
& Difference & 19.94\% & -1.26\% & 22.58\% & 11.95\%  \\
& t-test     
  & \textbf{\boldmath $p=.000$} & $p=.587$   
  & \textbf{\boldmath $p=.000$} & \textbf{\boldmath$p=.000$}   \\
\bottomrule
\end{tabular}
\label{tab:t-test}
\end{table}

\textit{Results.}
While this prompting strategy outperforms sensitivity prompting among Micro F1 scores, it performs very similarly in terms of Macro F1 scores. Additionally, we see a statistically significant change in Micro and Macro F1 scores among binary questions and Macro F1 scores among categorical questions compared to default prompting (see Figure \ref{tab:t-test}). These results suggest that even when provided with user POS scores and sensitivity ratings, LLMs still have a limited understanding of human privacy expectations.

\subsubsection{Few-Shot Prompting}
In all of the tests described above, the LLMs are limited to information regarding human \textit{perspectives}, such as privacy orientation and concern for privacy in a given scenario. The LLMs are not, however, exposed to the \textit{actions} participants would prefer in any given scenario. Therefore, our final prompting strategy is few-shot prompting~\cite{brown2020language}. We provide an example of how a participant assesses a single scenario, including the participant's POS scores, the sensitivity rating they provide, and their robot response preferences. After providing this information to the LLMs, we then ask them to evaluate the remaining two scenarios that the same participant was provided.

\textit{Results.}
In Table~\ref{tab:t-test}, we see a statistically significant, large increase in Macro F1 scores among binary questions as well as Micro and Macro F1 scores among categorical questions. The Micro F1 scores for binary questions do not show a substantial nor statistically significant difference compared to default prompting. This result is expected, as both the default-prompted LLM and human responses primarily include privacy-preserving choices and already show high overall accuracy.

\newcolumntype{C}[1]{>{\centering\arraybackslash}p{#1}}
\newcolumntype{R}[1]{>{\raggedright\arraybackslash}p{#1}}

\subsection{Case Study: LLM Logic \& Reasoning}
To further examine whether LLMs appropriately reason about participant privacy perspectives and preferences, we performed qualitative case studies on our best-performing LLM's \textit{reasoning} outputs. 

\subsubsection{Procedure}
To begin, we selected our overall best-performing model across all evaluation conditions, Grok3. We then compared Grok's default and few-shot model results (see Table~\ref{tab:default_f1_binary_categorical_multi}) along with its rationales. For each comparison, we examined pairs of responses, based on the same scenario and participant responses for each, and analyzed the differences across four types of scenarios:
\begin{enumerate}
    \item Both the default and few-shot results align well with the participant responses.
    \item Both the default and few-shot results align poorly with participant responses.
    \item The default results align well with participant responses, while the few-shot results do not.
    \item The default results align poorly with participant responses, while the few-shot results align well.
\end{enumerate}

In total, we produced 32 samples (\textit{i.e.}, four pairs of samples across the four scenario types). We analyzed these cases to understand Grok's performance in more depth.

\subsubsection{Results}
Overall, we observe that Grok provides a reasonable rationale for its responses, including references to details such as POS scores, sensitivity ratings, and scenario-specific preferences of participants, as well as information about the scenarios themselves. Here, we describe direct examples of Grok's rationale relevant to its decision making.

\textit{Case 1.}
In the first case, wherein the default and few-shot responses perfectly align with those of participants, the model maintains a similar rationale between conditions and adjusts some responses based on the details in the provided scenario. For example, Grok explicitly reported that, while sensitivity ratings are similar in both cases, it modified its response to one of the provided questions: \textit{``The main deviation is in data retention (Question 3), where I predict a shift from `delete immediately' to `keep temporarily' due to the utility of the data (reminders) and the lower concern level.''} This explanation shows that Grok is capable of producing sound responses and reasoning based on user instructions.

\textit{Case 2.} 
We see a similar pattern in the second case, but neither the default nor the few-shot responses match the human responses perfectly. This happens because, while the user chooses the same sensitivity score for both the example and the targeting scenario, they also select fewer privacy-enhancing choices in response to the targeting scenario. In this case, while Grok's reasoning is sensible, it has difficulty in predicting the participant's specific preferences. 

\textit{Case 3.}
In the third case, Grok's default results align more closely with the participant's responses than its few-shot results. When Grok provides default responses, it happens to select options that overall align well with the participant. In one example, Grok explains that privacy-enhancing options are ideal for the scenario, noting specific scenario details such as the involvement of a minor who \textit{``may not fully grasp the implications of data collection''}. Grok also makes logical concessions such as preferring that collected data be retained for a predefined period, rather than no data being retained, because the user in the scenario requested private data recording that must be kept to serve its purpose.

Once Grok receives additional information through the few-shot prompt, it is able to compare the participant's responses in the additional example scenario, and then proceeds to over-correct when responding to the current scenario. While Grok previously provided a rationale entirely based on the scenario's details, it instead focuses its responses on meeting the participant's preferences. For example, Grok specifically noted that it was attempting to balance both the high POS scores of the participant and their low scenario sensitivity rating. These differences, while entirely valid, appear to be difficult for Grok to mediate. As a result, Grok selected responses that were too relaxed, more consistent with a low sensitivity rating, and ultimately aligned poorly with the participant's preferences.

\textit{Case 4.}
In the fourth scenario, Grok's performance improves when provided with additional information (\textit{i.e.}, the few-shot prompt) compared to its default responses. By default, Grok again conveys that privacy-enhancing options are ideal, citing details like the involvement of a minor or the fact that the collected data is irrelevant to the robot's task. While these choices and rationales are reasonable, they do not align with the responses of the participant.

When Grok is provided with additional information, it then focuses its reasoning on the participant's preferences. For example, in one scenario, Grok stated that a participant's \textit{``moderate score on privacy as a right ... indicate[d] they may not always demand strict privacy measures if the context seems low-risk.''} In conjunction with the participant's low sensitivity rating of the scenario and relaxed privacy preferences in a scenario Grok deemed more sensitive (\textit{i.e.}, the scenario involved medical data), Grok chose responses that were more relaxed overall. As a result, Grok's responses were in greater alignment with the participant than they had been previously when selected agnostic of the participant.

\subsubsection{Patterns Across Scenarios}
Among Grok's rationales, we see frequent privacy-preserving responses under default prompting. When provided with details about participants, we see that Grok tends to mirror the responses from the example scenario during the targeting scenario. We calculated the percentage of scenarios in which this outcome occurs; Grok provides the exact same responses (\textit{i.e.}, both categorical and binary) in 41.9\% of scenarios. Among only binary responses, the percentage rises to 89.8\%.

While we observe that Grok provides identical responses to many of the examples it is provided, it acknowledges that these decisions are the same. It frequently refers to the preferences a participant previously expressed and purposefully suggests a similar response. A full analysis of Grok's logic across all scenarios, participants, and few-shot responses would be required to truly assess whether Grok is providing privacy-aware answers, but that analysis falls outside the scope of this work. These case studies, however, do suggest that Grok is utilizing provided details to generate appropriate and preferable responses to the particular user in question.

\section{Discussion}
In this work, we explored the relationship between privacy orientation and preferred robot behaviors, compared human-preferred robot behaviors to those provided by LLMs, and further evaluated various prompting strategies in terms of their ability to improve LLM performance. Regarding privacy orientation, we show that specific subscales of the POS are predictive of preferences for specific robot behaviors. Through our comparative analysis of human-preferred and LLM-generated robot responses, we show a low to moderate level of conformance (\textit{i.e.}, Micro and Macro F1 scores). In our exploration of prompting strategies, we saw a substantial increase in LLM performance through few-shot prompting, with more modest or no performance gains among other strategies. Through these results, we identify three broad implications of this work.

\subsection{Privacy Mindset}
The results of our user evaluation suggest that participants were highly privacy oriented (\textit{i.e.}, on average across all POS subscales) and preferred robots to behave in a privacy-preserving manner, even in minimally sensitive private scenarios. The difference between the POS scores of our participants and those of Baruh and Cemalc{\i}lar~\cite{baruh2014more} suggests a potential shift in privacy concerns since the original data collection in 2013. Given our sample was geographically based in the United States, we suspect that several major national events that have occurred since 2013 may have altered the privacy sentiments of our US-based participants. These include the government surveillance revelations of Edward Snowden in 2013 \cite{greenwald_et_al_2013_snowden}, concerns of user manipulation and election interference by Cambridge Analytica in 2016 \cite{detrow2018did}, and the Equifax data breach in 2017 \cite{epic_equifax_breach}, among others. Events of this magnitude could be responsible for an erosion in public trust of government and corporate agencies that collect large amounts of data and have significant responsibility for the management of that data. While we cannot draw any direct conclusions about these events, it is quite clear that the general U.S. population has concerns about privacy. Prior work by McClain et al.~\cite{pew2023privacy} has shown that more people in 2023 say they are not knowledgeable about what happens with their data and have little understanding of the laws and regulations that protect their privacy than compared to 2019. As users are continually required to make concessions about their privacy in order to use various technologies, it would be unsurprising if such a shift has occurred. As the scope of our study cannot identify evidence of an increase in privacy concerns broadly, it is also possible that our sample population is either skewed or biased by elements of our survey (\textit{e.g.}, introducing robot involvement in the survey prior to completion of the POS).

\subsection{LLM Privacy Awareness}
Our results show that LLMs generally provide privacy-preserving responses, but they are not necessarily privacy-aware. We observe a trend where LLMs provide increasingly private responses when they are provided higher sensitivity ratings; however, they often fail to capture the nuance of why a given scenario is sensitive or the appropriate response. From our case studies (see Section~\ref{sec:llm_eval}), we see that even the best-performing model can `overcorrect' during few-shot prompting when the details of one scenario lead the model to incorrectly infer stringent or weak privacy preferences in another. We also find that larger models, particularly the closed-source ones, perform better than open-source models. A cursory examination of the rationale of a small model that can be run on edge devices (\textit{i.e.}, Mistral), suggests that such models may not be able to provide valid privacy-related reasoning based on user inputs. Llama 4 Scout, however, performs similarly to larger models, potentially indicating a bottom line of model size for privacy-preserving rationale.

It is important to consider, however, that aligning privacy-related decisions to the unique preferences of another person can be a challenging task even for humans. Given the complexity of CI in privacy, it may be possible that humans would perform similarly to LLMs when attempting to align with peers. With that in mind, we cannot simply conclude that LLMs perform poorly at this task. Further research efforts will be needed to understand this comparison and to better contextualize LLM performance.

\subsection{LLM-Powered Robot Design Implications}
Prior works have already implemented LLMs and VLMs in robots directly~\cite{benjdira2023rosgpt_vision} or provided them with specific personas or rules~\cite{abbo2025can, bezabih2024toward}. In our study, however, we see that prompting LLMs with the addition of user privacy orientation or perceived sensitivity of scenarios only minimally improves performance, if not impairing performance or creating no change. Conversely, using few-shot prompting substantially improves LLM performance on average. These results imply that an LLM-powered robot can utilize naturalistic interactions with a given user to inform future privacy-sensitive responses to that same user.

\textit{Practical Considerations of LLM Decision Making.}
While few-shot prompting may be able to improve model conformance to user preferences, it is worth revisiting the methods used in our few-shot prompting strategy. In our work, due to practical limitations and the necessity of standardized inputs in our study design, we utilized a set of curated behaviors as responses to our example scenarios. In live scenarios, however, a robot will not have access to standardized multiple-choice behavior preferences. Instead, the robot will need to be capable of understanding ongoing scenarios and the unique responses those scenarios may require. For example, VLMs may be used to allow a robot to recognize its environment, as many researchers have already shown~\cite{wake2025vlm, zhong2024improving, rahimi2025user}. A VLM-equipped robot can subsequently reason about its environment, including details relevant to privacy, and appropriately tailor its response to the context of a scenario.

Additionally, the robot must be able to signal its data collection to users. For instance, robots may utilize privacy-preserving behaviors by default (\textit{e.g.}, minimize data collection and retention) and display both sound and text indicators (\textit{e.g.}, ``video recording in progress'') on the screen when needed. Future user studies are needed in order to design these indicators. Once a robot is able to recognize a scenario and respond to it, the robot then needs to be capable of discerning user responses or preferences in order to conform to users over time. While this outcome could be achieved through a user-driven calibration phase, only learning over time would allow the robot to understand the unique nuances of individual and family privacy preferences and how such preferences may change over time. For future work, we aim to implement such a system in an existing robot platform and conduct in-lab or in-home evaluations.

\textit{Personalization \& Data Privacy Tradeoff.}
As discussed in Section \ref{sec:llm_eval}, closed-source models generally perform better than open-source ones. Closed-source models, however, run over cloud API and, while most companies such as OpenAI do not use API user data for training by default~\cite{openai_enterprise_privacy_2025}, these data are stored temperately on cloud servers for debugging purposes. Given that prior works~\cite{afroogh2024trust, lockey2024trust, chen2025systematic} show that users have low levels of trust toward AI models' data processing practices, using non-local models on robots might cause privacy concerns. The goal of designing LLM-powered robots should account for this type of concern, and we, therefore, propose two potential solutions. First, if non-local model API calls are used, users should be clearly informed where their data is going and how their data will be used. Second, given the clear benefits of local models for user privacy (\textit{i.e.}, reduced data privacy concerns), future studies can be used to fine-tune these models, or distill them from open source ones, such that they are sufficiently privacy aware to also mitigate interpersonal privacy concerns.

\subsection{Limitations}
There are three primary limitations presented in this work. First, participant POS scores were higher on average in our study than in prior work. It is not clear whether this result reflects a true population change in privacy orientation over the last decade, as described earlier, or a skew or bias in our sample (see Section~\ref{app:POS_stats}). While many other scales exist that center on privacy concerns related to technology (see Section \ref{sec:conceptual_approaches_to_privacy}), many of these focus on either online privacy, concerns of specific populations, or privacy broadly. We have not identified, however, scales that are specific to in-context human-robot interaction. Development of such a scale may be worth exploring further as future work.

Secondly, while we have evaluated the performance of several models, identified strengths and weaknesses of those models, and explored strategies to improve model performance through prompting, we cannot definitively conclude whether LLMs are sufficiently privacy-aware. To do so, further qualitative analysis of state-of-the-art models is required. For example, if we are able to analyze the logic of all models across all of our scenarios, we may be able to better identify the unique cases in which models succeed or fail to meet user privacy preferences. In doing so, we can then gain a better understanding of whether LLMs are truly capable of privacy-aware decision making.

Finally, while our user study captures the perspectives of many participants across a diverse pool of scenarios, the evaluation is limited to U.S. participants and hypothetical scenarios. Therefore, future work can examine in-person evaluations and the inclusion of multicultural perspectives.

\section{Conclusion}
In this work, we have evaluated the privacy awareness of ten state-of-the-art LLMs in the context of household human-robot interaction. Our results show that LLMs possess broad limitations in terms of their ability to conform to user privacy preferences, but the prompting strategies we have presented may be able to bridge this gap. While the considerations of user privacy are varied and complex, the results of this work suggest that LLMs have the potential to act as rational privacy controllers for social robots capable of enhancing user privacy.

\newpage

\bibliographystyle{ACM-Reference-Format}
\bibliography{main}

@String{Computing = "Computing" }

@String{Computer = "{IEEE} Computer" }

@String{Chelsea = "Chelsea" }

@String{Springer = "Springer-Verlag" }

@misc{openai_o3_2025,
  author       = {OpenAI},
  title        = {OpenAI o3 Model Documentation},
  year         = {2025},
  url          = {https://platform.openai.com/docs/models/o3},
  note         = {Accessed: 2025-05-18}
}

@misc{google_gemini_2.5_flash_2025,
  author       = {Google},
  title        = {Gemini 2.5 Flash Preview Model Documentation},
  year         = {2025},
  url          = {https://ai.google.dev/gemini-api/docs/models#gemini-2.5-flash-preview},
  note         = {Accessed: 2025-05-18}
}

@article{costa2025securing,
  title={Securing AI Agents with Information-Flow Control},
  author={Costa, Manuel and K{\"o}pf, Boris and Kolluri, Aashish and Paverd, Andrew and Russinovich, Mark and Salem, Ahmed and Tople, Shruti and Wutschitz, Lukas and Zanella-B{\'e}guelin, Santiago},
  journal={arXiv preprint arXiv:2505.23643},
  year={2025}
}

@inproceedings{greshake2023not,
  title={Not what you've signed up for: Compromising real-world llm-integrated applications with indirect prompt injection},
  author={Greshake, Kai and Abdelnabi, Sahar and Mishra, Shailesh and Endres, Christoph and Holz, Thorsten and Fritz, Mario},
  booktitle={Proceedings of the 16th ACM workshop on artificial intelligence and security},
  pages={79--90},
  year={2023}
}

@misc{google_gemini_2.5_pro_2025,
  author       = {Google},
  title        = {Gemini 2.5 Pro Preview (05-06) Model Documentation},
  year         = {2025},
  url          = {https://ai.google.dev/gemini-api/docs/models#gemini-2.5-pro-preview-05-06},
  note         = {Accessed: 2025-05-18}
}

@misc{openai_gpt4o_2025,
  author       = {OpenAI},
  title        = {OpenAI GPT-4o Model Documentation},
  year         = {2025},
  url          = {https://platform.openai.com/docs/models/gpt-4o},
  note         = {Accessed: 2025-05-18}
}

@misc{xai_grok_models_2025,
  author       = {xAI},
  title        = {Grok Models and Pricing Documentation},
  year         = {2025},
  url          = {https://docs.x.ai/docs/models?cluster=us-east-1#models-and-pricing},
  note         = {Accessed: 2025-05-18}
}

@misc{deepseek_api_pricing_2025,
  author       = {DeepSeek},
  title        = {DeepSeek API Models and Pricing Documentation},
  year         = {2025},
  url          = {https://api-docs.deepseek.com/quick_start/pricing},
  note         = {Accessed: 2025-05-18}
}

@inproceedings{murali2023improving,
  title={Improving multiparty interactions with a robot using large language models},
  author={Murali, Prasanth and Steenstra, Ian and Yun, Hye Sun and Shamekhi, Ameneh and Bickmore, Timothy},
  booktitle={Extended Abstracts of the 2023 CHI Conference on Human Factors in Computing Systems},
  pages={1--8},
  year={2023}
}

@inproceedings{Amos_2021, series={WWW ’21},
   title={Privacy Policies over Time: Curation and Analysis of a Million-Document Dataset},
   url={http://dx.doi.org/10.1145/3442381.3450048},
   DOI={10.1145/3442381.3450048},
   booktitle={Proceedings of the Web Conference 2021},
   publisher={ACM},
   author={Amos, Ryan and Acar, Gunes and Lucherini, Eli and Kshirsagar, Mihir and Narayanan, Arvind and Mayer, Jonathan},
   year={2021},
   month=apr, pages={2165–2176},
   collection={WWW ’21} }

@article{chang2025keep,
  title={Keep Security! Benchmarking Security Policy Preservation in Large Language Model Contexts Against Indirect Attacks in Question Answering},
  author={Chang, Hwan and Kim, Yumin and Jun, Yonghyun and Lee, Hwanhee},
  journal={arXiv preprint arXiv:2505.15805},
  year={2025}
}

@article{ahmad2020policyqa,
  title={PolicyQA: A reading comprehension dataset for privacy policies},
  author={Ahmad, Wasi Uddin and Chi, Jianfeng and Tian, Yuan and Chang, Kai-Wei},
  journal={arXiv preprint arXiv:2010.02557},
  year={2020}
}

@article{azeem2024llm,
  title={Llm-driven robots risk enacting discrimination, violence, and unlawful actions},
  author={Azeem, Rumaisa and Hundt, Andrew and Mansouri, Masoumeh and Brand{\~a}o, Martim},
  journal={arXiv preprint arXiv:2406.08824},
  year={2024}
}

@inproceedings{zhong2024improving,
  title={Improving Visual Perception of a Social Robot for Controlled and In-the-wild Human-robot Interaction},
  author={Zhong, Wangjie and Tian, Leimin and Le, Duy Tho and Rezatofighi, Hamid},
  booktitle={Companion of the 2024 ACM/IEEE International Conference on Human-Robot Interaction},
  pages={1199--1203},
  year={2024}
}

@article{rahimi2025user,
  title={User-vlm 360: Personalized vision language models with user-aware tuning for social human-robot interactions},
  author={Rahimi, Hamed and Bahaj, Adil and Abrini, Mouad and Khoramshahi, Mahdi and Ghogho, Mounir and Chetouani, Mohamed},
  journal={arXiv preprint arXiv:2502.10636},
  year={2025}
}

@article{wake2025vlm,
  title={VLM-driven Behavior Tree for Context-aware Task Planning},
  author={Wake, Naoki and Kanehira, Atsushi and Takamatsu, Jun and Sasabuchi, Kazuhiro and Ikeuchi, Katsushi},
  journal={arXiv preprint arXiv:2501.03968},
  year={2025}
}

@article{bezabih2024toward,
  title={Toward LLM-Powered Social Robots for Supporting Sensitive Disclosures of Stigmatized Health Conditions},
  author={Bezabih, Alemitu and Nourriz, Shadi and Smith, C},
  journal={arXiv preprint arXiv:2409.04508},
  year={2024}
}

@article{abbo2025can,
  title={" Can you be my mum?": Manipulating Social Robots in the Large Language Models Era},
  author={Abbo, Giulio Antonio and Desideri, Gloria and Belpaeme, Tony and Spitale, Micol},
  journal={arXiv preprint arXiv:2501.04633},
  year={2025}
}

@article{benjdira2023rosgpt_vision,
  title={ROSGPT\_Vision: Commanding Robots Using Only Language Models' Prompts},
  author={Benjdira, Bilel and Koubaa, Anis and Ali, Anas M},
  journal={arXiv preprint arXiv:2308.11236},
  year={2023}
}

@misc{glba,
  author = {{U.S. Congress}},
  title = {Gramm-Leach-Bliley Act of 1999 (GLBA)},
  year = {1999},
  howpublished = {Public Law 106-102},
  note = {URL: \url{https://www.govinfo.gov/content/pkg/PLAW-106publ102/pdf/PLAW-106publ102.pdf}}
}

@misc{irc6103,
  author = {{Internal Revenue Service}},
  title = {IRC Section 6103 - Confidentiality and Disclosure of Returns and Return Information},
  year = {1976},
  howpublished = {26 U.S.C. § 6103},
  note = {URL: \url{https://www.law.cornell.edu/uscode/text/26/6103}}
}

@misc{fre502,
  author = {{United States Courts}},
  title = {Federal Rules of Evidence, Rule 502. Attorney-Client Privilege and Work Product; Limitations on Waiver},
  year = {2008},
  howpublished = {Federal Rules of Evidence},
  note = {URL: \url{https://www.uscourts.gov/rules-policies/current-rules-practice-procedure/federal-rules-evidence}}
}

@misc{ferpa,
  author = {{U.S. Congress}},
  title = {Family Educational Rights and Privacy Act of 1974 (FERPA)},
  year = {1974},
  howpublished = {20 U.S.C. § 1232g},
  note = {URL: \url{https://www.govinfo.gov/content/pkg/USCODE-2011-title20/pdf/USCODE-2011-title20-chap31-subchapIII-part4-sec1232g.pdf}}
}

@misc{office2003summary,
  title={Summary of the HIPAA privacy rule},
  author={Office for Civil Rights},
  journal={HIPAA Compliance Assistance},
  year={2003}
}

@article{zhang2024multi,
    title={Multi-P $^{2}$ A: A Multi-perspective Benchmark on Privacy Assessment for Large Vision-Language Models},
  author={Zhang, Jie and Cao, Xiangkui and Han, Zhouyu and Shan, Shiguang and Chen, Xilin},
  journal={arXiv preprint arXiv:2412.19496},
  year={2024}
}

@inproceedings{lockey2024trust,
  title={Trust in Artificial Intelligence: A Critical Systematic Review},
  author={Lockey, Steve and Gillespie, Nicole and Morrill, Jake and Pool, Javad},
  booktitle={Academy of Management Proceedings},
  volume={2024},
  number={1},
  pages={14990},
  year={2024},
  organization={Academy of Management Valhalla, NY 10595}
}

@article{chen2025systematic,
  title={A Systematic Review of User Attitudes Toward GenAI: Influencing Factors and Industry Perspectives},
  author={Chen, Junjie and Xie, Wei and Xie, Qing and Hu, Anshu and Qiao, Yiran and Wan, Ruoyu and Liu, Yuhan},
  journal={Journal of Intelligence},
  year={2025},
  publisher={Multidisciplinary Digital Publishing Institute}
}

@article{afroogh2024trust,
  title={Trust in AI: progress, challenges, and future directions},
  author={Afroogh, Saleh and Akbari, Ali and Malone, Emmie and Kargar, Mohammadali and Alambeigi, Hananeh},
  journal={Humanities and Social Sciences Communications},
  volume={11},
  number={1},
  pages={1--30},
  year={2024},
  publisher={Palgrave}
}

@misc{openai_enterprise_privacy_2025,
  title        = {Enterprise privacy at OpenAI},
  author       = {{OpenAI}},
  year         = {2025},
  month        = jun,
  day          = {4},
  note         = {Updated June 4, 2025},
  howpublished = {\url{https://openai.com/enterprise-privacy/}},
  urldate      = {2025-07-04}
}

@article{brown2020language,
  title={Language models are few-shot learners},
  author={Brown, Tom and Mann, Benjamin and Ryder, Nick and Subbiah, Melanie and Kaplan, Jared D and Dhariwal, Prafulla and Neelakantan, Arvind and Shyam, Pranav and Sastry, Girish and Askell, Amanda and others},
  journal={Advances in neural information processing systems},
  volume={33},
  pages={1877--1901},
  year={2020}
}

@inproceedings{bagdasarian2024airgapagent,
  title={Airgapagent: Protecting privacy-conscious conversational agents},
  author={Bagdasarian, Eugene and Yi, Ren and Ghalebikesabi, Sahra and Kairouz, Peter and Gruteser, Marco and Oh, Sewoong and Balle, Borja and Ramage, Daniel},
  booktitle={Proceedings of the 2024 on ACM SIGSAC Conference on Computer and Communications Security},
  pages={3868--3882},
  year={2024}
}

@article{yi2025privacy,
  title={Privacy Reasoning in Ambiguous Contexts},
  author={Yi, Ren and Suciu, Octavian and Gascon, Adria and Meiklejohn, Sarah and Bagdasarian, Eugene and Gruteser, Marco},
  journal={arXiv preprint arXiv:2506.12241},
  year={2025}
}

@inproceedings{denning2009spotlight,
  title={A spotlight on security and privacy risks with future household robots: attacks and lessons},
  author={Denning, Tamara and Matuszek, Cynthia and Koscher, Karl and Smith, Joshua R and Kohno, Tadayoshi},
  booktitle={Proceedings of the 11th international conference on Ubiquitous computing},
  pages={105--114},
  year={2009}
}

@article{bell2025always,
  title={" Is it always watching? Is it always listening?" Exploring Contextual Privacy and Security Concerns Toward Domestic Social Robots},
  author={Bell, Henry and Kwesi, Jabari and Laabadli, Hiba and Emami-Naeini, Pardis},
  journal={arXiv preprint arXiv:2507.10786},
  year={2025}
}

@misc{fell2024insecure,
  author       = {Julian Fell},
  title        = {Insecure Deebot robot vacuums collect photos and audio to train AI},
  howpublished = {ABC News, Story Lab},
  year         = {2024},
  month        = oct,
  day          = {4},
  note         = {Accessed via ABC News website},
  url          = {https://www.abc.net.au/news/2024-10-05/robot-vacuum-deebot-ecovacs-photos-ai/104416632}
}

@article{Guo2022RoombaPrivacy,
  author={Eileen Guo},
  title={A Roomba recorded a woman on the toilet. How did screenshots end up on Facebook?},
  journal={MIT Technology Review},
  year = {2022},
  url={https://www.technologyreview.com/2022/12/19/1065306/roomba-irobot-robot-vacuums-artificial-intelligence-training-data-privacy/}
}

@misc{grandview2025consumer,
  author       = {Grand View Research, Inc.},
  title        = {Consumer Robotics Market Size, Share \& Trends Analysis Report By Mode, Connectivity, Product, Region, and Segment Forecasts, 2025-2030},
  year         = {2024},
  url = {https://www.grandviewresearch.com/industry-analysis/consumer-robotics-market-report}
}

@misc{AIST_2013_Paro,
  author       = {National Institute of Advanced Industrial Science and Technology (AIST)},
  title        = {{AIST Story No.3: PARO the therapeutic robot seal}},
  year         = {2013},
  month        = {—},
  note         = {“Shibata says that more than 3000 PARO robots have already been introduced in around 30 countries.”},
  url          = {https://www.aist.go.jp/pdf/aist_e/aist_story/en_aist_story_no3.pdf},
  urldate      = {2025-07-03}
}

@article{Jack2024_PARIS,
  author       = {Louise Jack},
  title        = {Socially assistive robots deployed in Paris hospital to ease pressure on staff and reassure patients},
  journal      = {The National Robotarium (News)},
  year         = {2024},
  month        = {January},
  day          = {31},
  url          = {https://thenationalrobotarium.com/socially-assistive-robots-deployed-in-paris-hospital-to-ease-pressure-on-staff-and-reassure-patients/},
  note         = {Accessed via web search results; published January 31, 2024} 
}

@article{TheSun2025_NuraBot,
  author       = {{The Sun Tech}},
  title        = {AI robot nurse with creepy 'face' taking over hospital jobs as it patrols halls, delivers meds and tracks patient vitals},
  journal      = {The Sun},
  year         = {2025},
  month        = {May},
  day          = {21},
  url          = {https://www.the-sun.com/tech/14289670/ai-robot-nurse-hospital-video-nvidia/},

}

@misc{nysofa_elliq_initiative_2025,
  author       = {{New York State Office for the Aging}},
  title        = {ElliQ Proactive Care Companion Initiative},
  howpublished = {\url{https://aging.ny.gov/elliq-proactive-care-companion-initiative}},
  institution  = {New York State Office for the Aging},
  year         = {2025},
  note         = {Accessed: 2025-07-03},
}

@article{shao2024privacylens,
  title={Privacylens: Evaluating privacy norm awareness of language models in action},
  author={Shao, Yijia and Li, Tianshi and Shi, Weiyan and Liu, Yanchen and Yang, Diyi},
  journal={arXiv preprint arXiv:2409.00138},
  year={2024}
}

@article{zhu2024privauditor,
  title={PrivAuditor: Benchmarking Data Protection Vulnerabilities in LLM Adaptation Techniques},
  author={Zhu, Derui and Chen, Dingfan and Wu, Xiongfei and Geng, Jiahui and Li, Zhuo and Grossklags, Jens and Ma, Lei},
  journal={Advances in Neural Information Processing Systems},
  volume={37},
  pages={9668--9689},
  year={2024}
}

@misc{zharmagambetov2025agent,
      title={AgentDAM: Privacy Leakage Evaluation for Autonomous Web Agents}, 
      author={Arman Zharmagambetov and Chuan Guo and Ivan Evtimov and Maya Pavlova and Ruslan Salakhutdinov and Kamalika Chaudhuri},
      year={2025},
      eprint={2503.09780},
      archivePrefix={arXiv},
      primaryClass={cs.AI},
      url={https://arxiv.org/abs/2503.09780}, 
}

@article{ravichander2019question,
  title={Question answering for privacy policies: Combining computational and legal perspectives},
  author={Ravichander, Abhilasha and Black, Alan W and Wilson, Shomir and Norton, Thomas and Sadeh, Norman},
  journal={arXiv preprint arXiv:1911.00841},
  year={2019}
}

@article{sadeh2013usable,
  title={The usable privacy policy project},
  author={Sadeh, Norman and Acquisti, Alessandro and Breaux, Travis D and Cranor, Lorrie Faith and McDonald, Aleecia M and Reidenberg, Joel R and Smith, Noah A and Liu, Fei and Russell, N Cameron and Schaub, Florian and others},
  journal={Machine Learning and Natural Language Processing to Semi-Automatically Answer Those Privacy Questions Users Care About},
  pages={1--24},
  year={2013}
}

@article{zhang2024multitrust,
  title={Multitrust: A comprehensive benchmark towards trustworthy multimodal large language models},
  author={Zhang, Yichi and Huang, Yao and Sun, Yitong and Liu, Chang and Zhao, Zhe and Fang, Zhengwei and Wang, Yifan and Chen, Huanran and Yang, Xiao and Wei, Xingxing and others},
  journal={Advances in Neural Information Processing Systems},
  volume={37},
  pages={49279--49383},
  year={2024}
}

@article{gu2024mllmguard,
  title={Mllmguard: A multi-dimensional safety evaluation suite for multimodal large language models},
  author={Gu, Tianle and Zhou, Zeyang and Huang, Kexin and Dandan, Liang and Wang, Yixu and Zhao, Haiquan and Yao, Yuanqi and Yang, Yujiu and Teng, Yan and Qiao, Yu and others},
  journal={Advances in Neural Information Processing Systems},
  volume={37},
  pages={7256--7295},
  year={2024}
}

@misc{chi2023plue,
      title={PLUE: Language Understanding Evaluation Benchmark for Privacy Policies in English}, 
      author={Jianfeng Chi and Wasi Uddin Ahmad and Yuan Tian and Kai-Wei Chang},
      year={2023},
      eprint={2212.10011},
      archivePrefix={arXiv},
      primaryClass={cs.CL},
      url={https://arxiv.org/abs/2212.10011}, 
}

@article{hamid2023genaipabench,
  title={GenAIPABench: A benchmark for generative AI-based privacy assistants},
  author={Hamid, Aamir and Samidi, Hemanth Reddy and Finin, Tim and Pappachan, Primal and Yus, Roberto},
  journal={arXiv preprint arXiv:2309.05138},
  year={2023}
}

@article{mireshghallah2023can,
  title={Can llms keep a secret? testing privacy implications of language models via contextual integrity theory},
  author={Mireshghallah, Niloofar and Kim, Hyunwoo and Zhou, Xuhui and Tsvetkov, Yulia and Sap, Maarten and Shokri, Reza and Choi, Yejin},
  journal={arXiv preprint arXiv:2310.17884},
  year={2023}
}

@misc{dubois2025lengthcontrolledalpacaevalsimpleway,
      title={Length-Controlled AlpacaEval: A Simple Way to Debias Automatic Evaluators}, 
      author={Yann Dubois and Balázs Galambosi and Percy Liang and Tatsunori B. Hashimoto},
      year={2025},
      eprint={2404.04475},
      archivePrefix={arXiv},
      primaryClass={cs.LG},
      url={https://arxiv.org/abs/2404.04475}, 
}

@misc{zheng2023judging,
      title={Judging LLM-as-a-Judge with MT-Bench and Chatbot Arena}, 
      author={Lianmin Zheng and Wei-Lin Chiang and Ying Sheng and Siyuan Zhuang and Zhanghao Wu and Yonghao Zhuang and Zi Lin and Zhuohan Li and Dacheng Li and Eric P. Xing and Hao Zhang and Joseph E. Gonzalez and Ion Stoica},
      year={2023},
      eprint={2306.05685},
      archivePrefix={arXiv},
      primaryClass={cs.CL},
      url={https://arxiv.org/abs/2306.05685}, 
}

@article{austin2021program,
  title={Program synthesis with large language models},
  author={Austin, Jacob and Odena, Augustus and Nye, Maxwell and Bosma, Maarten and Michalewski, Henryk and Dohan, David and Jiang, Ellen and Cai, Carrie and Terry, Michael and Le, Quoc and others},
  journal={arXiv preprint arXiv:2108.07732},
  year={2021}
}

@article{srivastava2022beyond,
  title={Beyond the imitation game: Quantifying and extrapolating the capabilities of language models},
  author={Srivastava, Aarohi and Rastogi, Abhinav and Rao, Abhishek and Shoeb, Abu Awal Md and Abid, Abubakar and Fisch, Adam and Brown, Adam R and Santoro, Adam and Gupta, Aditya and Garriga-Alonso, Adri{\`a} and others},
  journal={arXiv preprint arXiv:2206.04615},
  year={2022}
}

@article{chen2021evaluating,
  title={Evaluating large language models trained on code},
  author={Chen, Mark and Tworek, Jerry and Jun, Heewoo and Yuan, Qiming and Pinto, Henrique Ponde De Oliveira and Kaplan, Jared and Edwards, Harri and Burda, Yuri and Joseph, Nicholas and Brockman, Greg and others},
  journal={arXiv preprint arXiv:2107.03374},
  year={2021}
}

@article{liang2020xglue,
  title={XGLUE: A new benchmark dataset for cross-lingual pre-training, understanding and generation},
  author={Liang, Yaobo and Duan, Nan and Gong, Yeyun and Wu, Ning and Guo, Fenfei and Qi, Weizhen and Gong, Ming and Shou, Linjun and Jiang, Daxin and Cao, Guihong and others},
  journal={arXiv preprint arXiv:2004.01401},
  year={2020}
}

@inproceedings{hu2020xtreme,
  title={Xtreme: A massively multilingual multi-task benchmark for evaluating cross-lingual generalisation},
  author={Hu, Junjie and Ruder, Sebastian and Siddhant, Aditya and Neubig, Graham and Firat, Orhan and Johnson, Melvin},
  booktitle={International conference on machine learning},
  pages={4411--4421},
  year={2020},
  organization={PMLR}
}

@article{zhang2023safetybench,
  title={SafetyBench: Evaluating the safety of large language models},
  author={Zhang, Zhexin and Lei, Leqi and Wu, Lindong and Sun, Rui and Huang, Yongkang and Long, Chong and Liu, Xiao and Lei, Xuanyu and Tang, Jie and Huang, Minlie},
  journal={arXiv preprint arXiv:2309.07045},
  year={2023}
}

@article{ruan2023identifying,
  title={Identifying the risks of lm agents with an lm-emulated sandbox},
  author={Ruan, Yangjun and Dong, Honghua and Wang, Andrew and Pitis, Silviu and Zhou, Yongchao and Ba, Jimmy and Dubois, Yann and Maddison, Chris J and Hashimoto, Tatsunori},
  journal={arXiv preprint arXiv:2309.15817},
  year={2023}
}

@article{fu2023practical,
  title={Practical membership inference attacks against fine-tuned large language models via self-prompt calibration},
  author={Fu, Wenjie and Wang, Huandong and Gao, Chen and Liu, Guanghua and Li, Yong and Jiang, Tao},
  journal={arXiv preprint arXiv:2311.06062},
  year={2023}
}

@article{mattern2023membership,
  title={Membership inference attacks against language models via neighbourhood comparison},
  author={Mattern, Justus and Mireshghallah, Fatemehsadat and Jin, Zhijing and Sch{\"o}lkopf, Bernhard and Sachan, Mrinmaya and Berg-Kirkpatrick, Taylor},
  journal={arXiv preprint arXiv:2305.18462},
  year={2023}
}

@article{zhuo2023red,
  title={Red teaming chatgpt via jailbreaking: Bias, robustness, reliability and toxicity},
  author={Zhuo, Terry Yue and Huang, Yujin and Chen, Chunyang and Xing, Zhenchang},
  journal={arXiv preprint arXiv:2301.12867},
  year={2023}
}

@article{staab2023beyond,
  title={Beyond memorization: Violating privacy via inference with large language models},
  author={Staab, Robin and Vero, Mark and Balunovi{\'c}, Mislav and Vechev, Martin},
  journal={arXiv preprint arXiv:2310.07298},
  year={2023}
}

@article{duan2024membership,
  title={Do membership inference attacks work on large language models?},
  author={Duan, Michael and Suri, Anshuman and Mireshghallah, Niloofar and Min, Sewon and Shi, Weijia and Zettlemoyer, Luke and Tsvetkov, Yulia and Choi, Yejin and Evans, David and Hajishirzi, Hannaneh},
  journal={arXiv preprint arXiv:2402.07841},
  year={2024}
}

@article{kim2023propile,
  title={Propile: Probing privacy leakage in large language models},
  author={Kim, Siwon and Yun, Sangdoo and Lee, Hwaran and Gubri, Martin and Yoon, Sungroh and Oh, Seong Joon},
  journal={Advances in Neural Information Processing Systems},
  volume={36},
  pages={20750--20762},
  year={2023}
}

@inproceedings{carlini2021extracting,
  title={Extracting training data from large language models},
  author={Carlini, Nicholas and Tramer, Florian and Wallace, Eric and Jagielski, Matthew and Herbert-Voss, Ariel and Lee, Katherine and Roberts, Adam and Brown, Tom and Song, Dawn and Erlingsson, Ulfar and others},
  booktitle={30th USENIX security symposium (USENIX Security 21)},
  pages={2633--2650},
  year={2021}
}

@article{bjorck2025gr00t,
  title={Gr00t n1: An open foundation model for generalist humanoid robots},
  author={Bjorck, Johan and Casta{\~n}eda, Fernando and Cherniadev, Nikita and Da, Xingye and Ding, Runyu and Fan, Linxi and Fang, Yu and Fox, Dieter and Hu, Fengyuan and Huang, Spencer and others},
  journal={arXiv preprint arXiv:2503.14734},
  year={2025}
}

@article{brohan2023rt,
  title={Rt-2: Vision-language-action models transfer web knowledge to robotic control},
  author={Brohan, Anthony and Brown, Noah and Carbajal, Justice and Chebotar, Yevgen and Chen, Xi and Choromanski, Krzysztof and Ding, Tianli and Driess, Danny and Dubey, Avinava and Finn, Chelsea and others},
  journal={arXiv preprint arXiv:2307.15818},
  year={2023}
}

@misc{meta_llama4_scout_2025,
  author       = {Meta Platforms, Inc.},
  title        = {Llama-4-Scout-17B-16E-Instruct},
  year         = {2025},
  url          = {https://huggingface.co/meta-llama/Llama-4-Scout-17B-16E-Instruct},
  note         = {Accessed: 2025-05-18}
}

@article{ahn2022can,
  title={Do as i can, not as i say: Grounding language in robotic affordances},
  author={Ahn, Michael and Brohan, Anthony and Brown, Noah and Chebotar, Yevgen and Cortes, Omar and David, Byron and Finn, Chelsea and Fu, Chuyuan and Gopalakrishnan, Keerthana and Hausman, Karol and others},
  journal={arXiv preprint arXiv:2204.01691},
  year={2022}
}

@article{hassan2024integrating,
  title={Integrating Vision and Olfaction via Multi-Modal LLM for Robotic Odor Source Localization},
  author={Hassan, Sunzid and Wang, Lingxiao and Mahmud, Khan Raqib},
  journal={Sensors},
  volume={24},
  number={24},
  pages={7875},
  year={2024},
  publisher={MDPI}
}

@inproceedings{yuan2024towards,
  title={Towards an llm-based speech interface for robot-assisted feeding},
  author={Yuan, Jessie and Gupta, Janavi and Padmanabha, Akhil and Karachiwalla, Zulekha and Majidi, Carmel and Admoni, Henny and Erickson, Zackory},
  booktitle={Adjunct Proceedings of the 37th Annual ACM Symposium on User Interface Software and Technology},
  pages={1--4},
  year={2024}
}

@article{shinn2023reflexion,
  title={Reflexion: Language agents with verbal reinforcement learning},
  author={Shinn, Noah and Cassano, Federico and Gopinath, Ashwin and Narasimhan, Karthik and Yao, Shunyu},
  journal={Advances in Neural Information Processing Systems},
  volume={36},
  pages={8634--8652},
  year={2023}
}

@inproceedings{liang2023code,
  title={Code as policies: Language model programs for embodied control},
  author={Liang, Jacky and Huang, Wenlong and Xia, Fei and Xu, Peng and Hausman, Karol and Ichter, Brian and Florence, Pete and Zeng, Andy},
  booktitle={2023 IEEE International Conference on Robotics and Automation (ICRA)},
  pages={9493--9500},
  year={2023},
  organization={IEEE}
}

@misc{mistral2024mistral7b,
  author       = {Mistral AI},
  title        = {Mistral-7B-Instruct-v0.3},
  year         = {2024},
  howpublished = {\url{https://huggingface.co/mistralai/Mistral-7B-Instruct-v0.3}},
  note         = {Released May 22, 2024. Licensed under Apache 2.0.}
}

@misc{meta2024llama33,
  author       = {Meta Platforms, Inc.},
  title        = {Llama 3.3 70B Instruct},
  year         = {2024},
  howpublished = {\url{https://huggingface.co/meta-llama/Llama-3.3-70B-Instruct}},
  note         = {Released December 6, 2024. Licensed under the Llama 3.3 Community License Agreement.}
}

@article{nissenbaum2004privacy,
  title={Privacy as contextual integrity},
  author={Nissenbaum, Helen},
  journal={Wash. L. Rev.},
  volume={79},
  pages={119},
  year={2004},
  publisher={HeinOnline}
}

@article{yang2025qwen3,
  title={Qwen3 Technical Report},
  author={Yang, An and Li, Anfeng and Yang, Baosong and Zhang, Beichen and Hui, Binyuan and Zheng, Bo and Yu, Bowen and Gao, Chang and Huang, Chengen and Lv, Chenxu and others},
  journal={arXiv preprint arXiv:2505.09388},
  year={2025}
}

@article{baruh2014more,
  title={It is more than personal: Development and validation of a multidimensional privacy orientation scale},
  author={Baruh, Lemi and Cemalc{\i}lar, Zeynep},
  journal={Personality and Individual Differences},
  volume={70},
  pages={165--170},
  year={2014},
  publisher={Elsevier}
}

@article{brause2024there,
  title={‘There are some things that I would never ask Alexa’--privacy work, contextual integrity, and smart speaker assistants},
  author={Brause, Saba Rebecca and Blank, Grant},
  journal={Information, Communication \& Society},
  volume={27},
  number={1},
  pages={182--197},
  year={2024},
  publisher={Taylor \& Francis}
}

@incollection{nissenbaum2009privacy,
  title={Privacy in context: Technology, policy, and the integrity of social life},
  author={Nissenbaum, Helen},
  booktitle={Privacy in context},
  year={2009},
  publisher={Stanford University Press}
}

@article{preibusch2013guide,
  title={Guide to measuring privacy concern: Review of survey and observational instruments},
  author={Preibusch, S{\"o}ren},
  journal={International journal of human-computer studies},
  volume={71},
  number={12},
  pages={1133--1143},
  year={2013},
  publisher={Elsevier}
}

@article{elueze2018privacy,
  title={Privacy attitudes and concerns in the digital lives of older adults: Westin’s privacy attitude typology revisited},
  author={Elueze, Isioma and Quan-Haase, Anabel},
  journal={American Behavioral Scientist},
  volume={62},
  number={10},
  pages={1372--1391},
  year={2018},
  publisher={SAGE Publications Sage CA: Los Angeles, CA}
}

@inproceedings{chignell2003privacy,
  title={The privacy attitudes questionnaire (paq): initial development and validation},
  author={Chignell, Mark H and Quan-Haase, Anabel and Gwizdka, Jacek},
  booktitle={proceedings of the human factors and ergonomics society annual meeting},
  volume={47},
  number={11},
  pages={1326--1330},
  year={2003},
  organization={SAGE Publications Sage CA: Los Angeles, CA}
}

@inproceedings{giang2023privacy,
  title={The Privacy Attitudes Questionnaire (PAQ) Turned 20: What’s Next for a Generalized Measure of Privacy Attitudes?},
  author={Giang, Wayne CW and Iglar, Alyssa and Chignell, Mark and Samavi, Reza and Nathan, Kopiha},
  booktitle={Proceedings of the Human Factors and Ergonomics Society Annual Meeting},
  volume={67},
  number={1},
  pages={705--709},
  year={2023},
  organization={SAGE Publications Sage CA: Los Angeles, CA}
}

@article{hrynenko2024identifying,
  title={Identifying Privacy Personas},
  author={Hrynenko, Olena and Cavallaro, Andrea},
  journal={arXiv preprint arXiv:2410.14023},
  year={2024}
}

@inproceedings{farzand2024out,
  title={Out-of-device privacy unveiled: Designing and validating the out-of-device privacy scale (odps)},
  author={Farzand, Habiba and Marky, Karola and Khamis, Mohamed},
  booktitle={Proceedings of the 2024 CHI Conference on Human Factors in Computing Systems},
  pages={1--15},
  year={2024}
}

@article{biselli2022challenges,
  title={On the challenges of developing a concise questionnaire to identify privacy personas},
  author={Biselli, Tom and Steinbrink, Enno and Herbert, Franziska and Schmidbauer-Wolf, Gina M and Reuter, Christian},
  journal={Proceedings on Privacy Enhancing Technologies},
  year={2022}
}

@article{buchanan2007development,
  title={Development of measures of online privacy concern and protection for use on the Internet},
  author={Buchanan, Tom and Paine, Carina and Joinson, Adam N and Reips, Ulf-Dietrich},
  journal={Journal of the American society for information science and technology},
  volume={58},
  number={2},
  pages={157--165},
  year={2007},
  publisher={Wiley Online Library}
}

@article{guerrero2017cybersecurity,
  title={Cybersecurity of robotics and autonomous systems: Privacy and safety},
  author={Guerrero, {\'A}ngel Manuel},
  journal={Robotics: legal, ethical and socioeconomic impacts},
  pages={75},
  year={2017},
  publisher={BoD--Books on Demand}
}

@inproceedings{rueben2018themes,
  title={Themes and research directions in privacy-sensitive robotics},
  author={Rueben, Matthew and Aroyo, Alexander Mois and Lutz, Christoph and Schm{\"o}lz, Johannes and Van Cleynenbreugel, Pieter and Corti, Andrea and Agrawal, Siddharth and Smart, William D},
  booktitle={2018 IEEE workshop on advanced robotics and its social impacts (ARSO)},
  pages={77--84},
  year={2018},
  organization={IEEE}
}

@article{torresen2018review,
  title={A review of future and ethical perspectives of robotics and AI},
  author={Torresen, Jim},
  journal={Frontiers in Robotics and AI},
  volume={4},
  pages={75},
  year={2018},
  publisher={Frontiers Media SA}
}

@article{pagallo2018rise,
  title={The rise of robotics \& AI: technological advances \& normative dilemmas},
  author={Pagallo, Ugo and Corrales, Marcelo and Fenwick, Mark and Forg{\'o}, Nikolaus},
  journal={Robotics, AI and the Future of Law},
  pages={1--13},
  year={2018},
  publisher={Springer}
}

@article{lutz2019privacy,
  title={The privacy implications of social robots: Scoping review and expert interviews},
  author={Lutz, Christoph and Sch{\"o}ttler, Maren and Hoffmann, Christian Pieter},
  journal={Mobile Media \& Communication},
  volume={7},
  number={3},
  pages={412--434},
  year={2019},
  publisher={SAGE Publications Sage UK: London, England}
}

@article{chatzimichali2020toward,
  title={Toward privacy-sensitive human--robot interaction: Privacy terms and human--data interaction in the personal robot era},
  author={Chatzimichali, Anna and Harrison, Ross and Chrysostomou, Dimitrios},
  journal={Paladyn, Journal of Behavioral Robotics},
  volume={12},
  number={1},
  pages={160--174},
  year={2020},
  publisher={De Gruyter}
}

@inproceedings{levinson2024snitches,
  title={Snitches Get Unplugged: Adolescents' Privacy Concerns about Robots in the Home are Relationally Situated},
  author={Levinson, Leigh and Nippert-Eng, Christena and Gomez, Randy and Sabanovi{\'c}, Selma},
  booktitle={Proceedings of the 2024 ACM/IEEE International Conference on Human-Robot Interaction},
  pages={423--432},
  year={2024}
}

@inproceedings{fernandes2016detection,
  title={Detection of privacy-sensitive situations for social robots in smart homes},
  author={Fernandes, Francisco Erivaldo and Yang, Guanci and Do, Ha Manh and Sheng, Weihua},
  booktitle={2016 IEEE International Conference on Automation Science and Engineering (CASE)},
  pages={727--732},
  year={2016},
  organization={IEEE}
}

@article{urquhart2019responsible,
  title={Responsible domestic robotics: exploring ethical implications of robots in the home},
  author={Urquhart, Lachlan and Reedman-Flint, Dominic and Leesakul, Natalie},
  journal={Journal of Information, Communication and Ethics in Society},
  volume={17},
  number={2},
  pages={246--272},
  year={2019},
  publisher={Emerald Publishing Limited}
}

@inproceedings{butler2015privacy,
  title={The privacy-utility tradeoff for remotely teleoperated robots},
  author={Butler, Daniel J and Huang, Justin and Roesner, Franziska and Cakmak, Maya},
  booktitle={Proceedings of the tenth annual ACM/IEEE international conference on human-robot interaction},
  pages={27--34},
  year={2015}
}

@inproceedings{leite2016robot,
  title={The robot who knew too much: Toward understanding the privacy/personalization trade-off in child-robot conversation},
  author={Leite, Iolanda and Lehman, Jill Fain},
  booktitle={Proceedings of the The 15th International Conference on Interaction Design and Children},
  pages={379--387},
  year={2016}
}

@article{lutz2020robot,
  title={The robot privacy paradox: Understanding how privacy concerns shape intentions to use social robots},
  author={Lutz, Christoph and Tam{\'o}-Larrieux, Aurelia},
  journal={Human-Machine Communication},
  volume={1},
  pages={87--111},
  year={2020},
  publisher={Communication and Social Robotics Labs Kalamazoo, Michigan}
}

@inproceedings{sullivan2025protecting,
  title={Protecting User Data Through Privacy-Sensitive Robot Design},
  author={Sullivan, Dakota and Mutlu, Bilge},
  booktitle={2025 20th ACM/IEEE International Conference on Human-Robot Interaction (HRI)},
  pages={1891--1893},
  year={2025},
  organization={IEEE}
}

@inproceedings{kim2019privacy,
  title={Privacy-preserving robot vision with anonymized faces by extreme low resolution},
  author={Kim, Myeung Un and Lee, Harim and Yang, Hyun Jong and Ryoo, Michael S},
  booktitle={2019 IEEE/RSJ International Conference on Intelligent Robots and Systems (IROS)},
  pages={462--467},
  year={2019},
  organization={IEEE}
}

@inproceedings{eick2020enhancing,
  title={Enhancing privacy in robotics via judicious sensor selection},
  author={Eick, Stephen and Ant{\'o}n, Annie I},
  booktitle={2020 IEEE International Conference on Robotics and Automation (ICRA)},
  pages={7156--7165},
  year={2020},
  organization={IEEE}
}

@inproceedings{shome2023robots,
  title={Robots as AI Double Agents: Privacy in Motion Planning},
  author={Shome, Rahul and Kingston, Zachary and Kavraki, Lydia E},
  booktitle={2023 IEEE/RSJ International Conference on Intelligent Robots and Systems (IROS)},
  pages={2861--2868},
  year={2023},
  organization={IEEE}
}

@article{vaswani2017attention,
  title={Attention is all you need},
  author={Vaswani, Ashish and Shazeer, Noam and Parmar, Niki and Uszkoreit, Jakob and Jones, Llion and Gomez, Aidan N and Kaiser, {\L}ukasz and Polosukhin, Illia},
  journal={Advances in neural information processing systems},
  volume={30},
  year={2017}
}

@article{radford2018improving,
  title={Improving language understanding by generative pre-training},
  author={Radford, Alec and Narasimhan, Karthik and Salimans, Tim and Sutskever, Ilya and others},
  year={2018},
  publisher={San Francisco, CA, USA}
}

@article{mai2023llm,
  title={Llm as a robotic brain: Unifying egocentric memory and control},
  author={Mai, Jinjie and Chen, Jun and Qian, Guocheng and Elhoseiny, Mohamed and Ghanem, Bernard and others},
  year={2023},
  publisher={arXiv}
}

@article{yoshikawa2023large,
  title={Large language models for chemistry robotics},
  author={Yoshikawa, Naruki and Skreta, Marta and Darvish, Kourosh and Arellano-Rubach, Sebastian and Ji, Zhi and Bj{\o}rn Kristensen, Lasse and Li, Andrew Zou and Zhao, Yuchi and Xu, Haoping and Kuramshin, Artur and others},
  journal={Autonomous Robots},
  volume={47},
  number={8},
  pages={1057--1086},
  year={2023},
  publisher={Springer}
}

@article{kim2024framework,
    title={Framework for Integrating Large Language Models with a Robotic Health Attendant for Adaptive Task Execution in Patient Care},
    author={Kim, Kyungki and Windle, John and Christian, Melissa and Windle, Tom and Ryherd, Erica and Huang, Pei-Chi and Robinson, Anthony and Chapman, Reid},
    journal={Applied Sciences},
    volume={14},
    number={21},
    pages={},
    year={2024},
    doi={10.3390/app14219922},
    url={https://www.mdpi.com/2076-3417/14/21/9922}
}

@inproceedings{kim2024understanding,
    author = {Kim, Callie Y. and Lee, Christine P. and Mutlu, Bilge},
    title = {Understanding Large-Language Model (LLM)-powered Human-Robot Interaction},
    year = {2024},
    isbn = {9798400703225},
    publisher = {Association for Computing Machinery},
    address = {New York, NY, USA},
    url = {https://doi.org/10.1145/3610977.3634966},
    doi = {10.1145/3610977.3634966},
    booktitle = {Proceedings of the 2024 ACM/IEEE International Conference on Human-Robot Interaction},
    pages = {371–380},
    numpages = {10},
    location = {Boulder, CO, USA},
    series = {HRI '24}
}

@misc{zeng2023large,
      title={Large Language Models for Robotics: A Survey}, 
      author={Fanlong Zeng and Wensheng Gan and Yongheng Wang and Ning Liu and Philip S. Yu},
      year={2023},
      eprint={2311.07226},
      archivePrefix={arXiv},
      primaryClass={cs.RO},
      url={https://arxiv.org/abs/2311.07226}, 
}

@misc{irobot2024financial,
  author       = {iRobot Corporation},
  title        = {iRobot Reports Fourth-Quarter and Full-Year 2024 Financial Results},
  year         = {2025},
  url          = {https://investor.irobot.com/news-releases/news-release-details/irobot-reports-fourth-quarter-and-full-year-2024-financial}
}

@inproceedings{choe2011living,
author = {Choe, Eun Kyoung and Consolvo, Sunny and Jung, Jaeyeon and Harrison, Beverly and Kientz, Julie A.},
title = {Living in a glass house: a survey of private moments in the home},
year = {2011},
isbn = {9781450306300},
publisher = {Association for Computing Machinery},
address = {New York, NY, USA},
url = {https://doi.org/10.1145/2030112.2030118},
doi = {10.1145/2030112.2030118},
booktitle = {Proceedings of the 13th International Conference on Ubiquitous Computing},
pages = {41–44},
numpages = {4},
location = {Beijing, China},
series = {UbiComp '11}
}

@incollection{clarke2014thematic,
  title={Thematic analysis},
  author={Clarke, Victoria and Braun, Virginia},
  booktitle={Encyclopedia of critical psychology},
  pages={1947--1952},
  year={2014},
  publisher={Springer}
}

@article{detrow2018did,
  title={What did Cambridge Analytica do during the 2016 election},
  author={Detrow, Scott},
  journal={National Public Radio},
  volume={20},
  year={2018}
}

@misc{epic_equifax_breach,
  author       = {{Electronic Privacy Information Center (EPIC)}},
  title        = {Equifax Data Breach},
  howpublished = {\url{https://archive.epic.org/privacy/data-breach/equifax/}},
  note         = {Accessed July 11, 2025},
  year         = {2017},
  month        = {sep},
}

@article{greenwald_et_al_2013_snowden,
  author       = {Glenn Greenwald and Ewen MacAskill and Laura Poitras},
  title        = {Edward Snowden: the whistleblower behind the {NSA} surveillance revelations},
  journal      = {The Guardian},
  year         = {2013},
  url          = {https://www.theguardian.com/world/2013/jun/09/edward-snowden-nsa-whistleblower-surveillance},
}

@misc{pew2023privacy,
  author       = {McClain, Colleen and Faverio, Michelle and Anderson, Monica and Park, Eugenie},
  title        = {Views of Data Privacy Risks, Personal Data and Digital Privacy Laws},
  year         = {2023},
  url          ={https://www.pewresearch.org/internet/2023/10/18/views-of-data-privacy-risks-personal-data-and-digital-privacy-laws/}
}

@article{levinson2024our,
  title={Our business, not the robot’s: family conversations about privacy with social robots in the home},
  author={Levinson, Leigh and McKinney, Jessica and Nippert-Eng, Christena and Gomez, Randy and {\v{S}}abanovi{\'c}, Selma},
  journal={Frontiers in Robotics and AI},
  volume={11},
  pages={1331347},
  year={2024},
  publisher={Frontiers Media SA}
}

@inproceedings{tang2022confidant,
  title={Confidant: A privacy controller for social robots},
  author={Tang, Brian and Sullivan, Dakota and Cagiltay, Bengisu and Chandrasekaran, Varun and Fawaz, Kassem and Mutlu, Bilge},
  booktitle={2022 17th ACM/IEEE International Conference on Human-Robot Interaction (HRI)},
  pages={205--214},
  year={2022},
  organization={IEEE}
}

@article{apthorpe2018discovering,
  title={Discovering smart home internet of things privacy norms using contextual integrity},
  author={Apthorpe, Noah and Shvartzshnaider, Yan and Mathur, Arunesh and Reisman, Dillon and Feamster, Nick},
  journal={Proceedings of the ACM on interactive, mobile, wearable and ubiquitous technologies},
  volume={2},
  number={2},
  pages={1--23},
  year={2018},
  publisher={ACM New York, NY, USA}
}

@article{hennink2022sample,
  title={Sample sizes for saturation in qualitative research: A systematic review of empirical tests},
  author={Hennink, Monique and Kaiser, Bonnie N},
  journal={Social science \& medicine},
  volume={292},
  pages={114523},
  year={2022},
  publisher={Elsevier}
}

@article{westfall2014statistical,
  title={Statistical power and optimal design in experiments in which samples of participants respond to samples of stimuli.},
  author={Westfall, Jacob and Kenny, David A and Judd, Charles M},
  journal={Journal of Experimental Psychology: General},
  volume={143},
  number={5},
  pages={2020},
  year={2014},
  publisher={American Psychological Association}
}


\clearpage 
\appendix

\section{Codebook}
\label{app:codebook}
\begin{table}[ht]
\caption{Robot Behavior Codes---The results of our second pre-study focused on identifying robot response behaviors following collection of private data. These behaviors broadly fit within four categories: data collection, user notification, data management, and data sharing.}
\scriptsize
  \begin{tabularx}{\columnwidth}{l X r}
    \toprule
    \textbf{Category} & \textbf{Code} & \textbf{Count} \\ 
    \toprule
    \multirow{2}{*}{Data Collection}
        & Do not record or discomfort with robot & 6 \\
        & Allow users to temporarily pause data collection & 2 \\
    \midrule
    \multirow{7}{*}{User Notification}
        & Notify user of data collection & 24 \\
        & Offer ability to review, manage, or delete data & 23 \\
        & Behave as normal or no change in behavior & 15 \\
        & Do not interfere with user & 10 \\
        & Ask user how to manage data & 8 \\
        & Be transparent about data collection & 7 \\
        & Explain purpose of collection & 6 \\
    \midrule
    \multirow{8}{*}{Data Management}
        & Securely store data & 30 \\
        & Delete data after a predefined period & 16 \\
        & Immediately delete data & 11 \\
        & Retain only essential data & 11 \\
        & Appraise data sensitivity & 7 \\
        & Learn or improve from data & 5 \\
        & Store data locally only & 5 \\
        & Anonymize data & 3\\
    \midrule
    \multirow{5}{*}{Data Sharing}
        & Do not share data with anyone except user & 13 \\
        & Only refer to data when user requests it & 6 \\
        & Only share or use data when given permission & 5 \\
        & Do not upload data online or with manufacturer without permission & 5\\
        & Share data only with family & 2\\
    \bottomrule
  \end{tabularx}
  
  \label{tab:codes-vertical}
\end{table}

\section{POS Statistics Comparing to Prior work}
  \label{app:POS_stats}
\begin{table}[ht]
  \caption{Comparative POS Statistics---Mean, standard deviation, and reliability POS statistics of our population and that of the original sample of \cite{baruh2014more}.}
\scriptsize
  \begin{tabular}{l l c c c r}
    \toprule
    \textbf{Sample} & \textbf{Statistic} & \textbf{Sub. 1} & \textbf{Sub. 2} & \textbf{Sub. 3} & \textbf{Sub. 4} \\ 
    \toprule
    \multirow{3}{*}{Our Sample} 
        & Mean & 4.49 & 3.84 & 3.96 & 4.50 \\ 
        & SD & 0.61 & 0.81 & 0.77 & 0.52\\ 
        & Reliability & 0.77 & 0.75 & 0.75 & 0.76 \\ 
    \midrule
    \multirow{3}{*}{Baruh \& Cemalcılar}
        & Mean & 4.09 & 3.37 & 3.48 & 4.08 \\ 
        & SD & 0.73 & 0.88 & 0.82 & 0.65\\ 
        & Reliability & 0.83 & 0.83 & 0.85 & 0.88 \\ 
    \bottomrule
  \end{tabular}
\end{table}
\newpage
\section{Household Patterns}
  \label{app:user-stats}
\begin{table}[ht]
\scriptsize
  \caption{Household Patterns---The results of our first pre-study focused on identifying the frequency of the following household patterns: the type of room in which the participant typically resides and the number, age, and relation of users present in the home.}
  \begin{tabular}{l l c c}
    \toprule
    \textbf{Variable} & \textbf{Category} & \textbf{Mean (\%)} & \textbf{SD} \\ 
    \toprule
    \multirow{2}{*}{Type of Location} 
        & Private & 48.94 & 24.34 \\ 
        & Shared  & 51.06 & 24.34 \\
    \midrule
    \multirow{2}{*}{Number of Users} 
        & Single & 37.22 & 34.97 \\ 
        & Multiple  & 62.78 & 34.97 \\
    \midrule
    \multirow{3}{*}{Age of Users} 
        & Children & 13.12 & 22.08 \\ 
        & Adults   & 52.96 & 40.32 \\ 
        & Both     & 33.92 & 37.15 \\
    \midrule
    \multirow{3}{*}{Relation of Users} 
        & Nuclear family   & 78.07 & 22.18 \\ 
        & Extended family  & 11.01 & 13.23 \\ 
        & Outsiders        & 10.92 & 16.13 \\ 
    \bottomrule
  \end{tabular}

\end{table}

\end{document}